\pdfoutput=1
\documentclass[11pt]{article}
\usepackage[final]{acl}
\usepackage{changepage}

\usepackage{times}
\usepackage{latexsym}
\usepackage[T1]{fontenc}
\usepackage[utf8]{inputenc}

\usepackage{microtype}
\usepackage{inconsolata}
\usepackage{graphicx}
\usepackage{tcolorbox}
\usepackage{amsmath}
\usepackage{subcaption}
\usepackage{url} 
\usepackage{booktabs}
\usepackage{amssymb}
\usepackage{todonotes}
\usepackage{multirow}
\usepackage{colortbl}
\usepackage{xcolor}

\usepackage{tcolorbox}  
\usepackage[dvipsnames]{xcolor}

\usepackage{tabularx}
\usepackage{array}
\usepackage{caption}

\usepackage{tikz}
\usetikzlibrary{arrows.meta, positioning, calc, decorations.pathreplacing}

\definecolor{mediumred}{RGB}{255,100,100}

\newcolumntype{Y}{>{\raggedright\arraybackslash}X}

\title{DFKI-MLT at SemEval-2026 TASK 7: Steering Multilingual Models Towards Cultural Knowledge}
\author{
  \textbf{Yusser Al Ghussin$^{1,2}$} \enspace \textbf{Daniil Gurgurov$^{1,2}$} \enspace \textbf{Yasser Hamidullah$^{1,2}$} \\ \textbf{Josef van Genabith$^{1,2}$} \enspace \textbf{Cristina España-Bonet$^{1,3}$} \enspace \textbf{Simon Ostermann$^{1,2}$} \\
  \\
  \quad \quad \quad  \quad$^{1}$German Research Center for Artificial Intelligence (DFKI GmbH),\\
\quad \quad \quad \quad $^{2}$Saarland Informatics Campus, Saarbrücken, Germany \\
\quad \quad \quad \quad $^{3}$Barcelona Supercomputing Center (BSC-CNS), Barcelona, Catalonia, Spain\\[6pt]
}

\begin{document}
\maketitle

\begin{abstract}

Large language models (LLMs) are increasingly used across diverse linguistic and cultural contexts, yet their cultural knowledge remains uneven across regions and languages. We present the \textbf{DFKI-MLT} system for \textbf{SemEval-2026 Task~7} on cultural awareness, where we apply \emph{activation steering} to multilingual LLMs using language vectors extracted from parallel FLORES data. Our method performs inference-time adaptation by adding language-specific steering vectors to the residual stream at a selected transformer layer, without any parameter updates.
We participated in both the short-answer (SAQ) and multiple-choice (MCQ) tracks; however, only our MCQ submission received an official score. In the official \textbf{MCQ} track, we achieved \textbf{86.96\%} accuracy, ranking \textbf{7th out of 17} teams.
To better understand system behavior, we conduct post-hoc analyses on the shared-task MCQ and SAQ settings. 
These analyses show that activation steering yields \emph{modest} and \emph{heterogeneous} improvements on cultural reasoning: gains are strongly \emph{layer-sensitive}, vary substantially across language--region pairs (some configurations even degrade performance), and interact with prompt formulation (generic vs.\ culturally conditioned prompts). Our findings suggest that prompt design and activation steering should be jointly optimized for culturally aware multilingual inference.
We release our code and experimental configurations at \url{https://github.com/Yusser96/SemEval-2026-Track7}.
\end{abstract}

\section{Introduction}

\begin{figure}[t]
    \centering
    \begin{tikzpicture}[>=Stealth, scale=0.75]
        
        \definecolor{langblue}{RGB}{65, 105, 225}
        \definecolor{cultred}{RGB}{220, 80, 80}
        \definecolor{specgreen}{RGB}{60, 160, 100}
        \definecolor{gridgray}{RGB}{230, 230, 230}
        
        \draw[gridgray, very thin] (0,0) grid[step=0.5] (4.0,3.5);
        
        \draw[->, thick, gray!70] (-0.3,0) -- (4,0);
        \draw[->, thick, gray!70] (0,-0.3) -- (0,4);
        
        \filldraw[black] (0,0) circle (2pt);
        
        \draw[-{Stealth[length=3mm, width=2mm]}, ultra thick, langblue] 
            (0,0) -- (1.8,2.6);
        
        \draw[-{Stealth[length=3mm, width=2mm]}, ultra thick, cultred] 
            (0,0) -- (1.4,3);
        
        
        
        \node[langblue, font=\small\sffamily, below] at (2.3,2.1) {Language};
        \node[cultred, font=\small\sffamily, above left] at (1.8,3.1) {Culture};
        
        \begin{scope}[shift={(5.0,1)}]
            \node[font=\small\sffamily\bfseries, anchor=west] at (0,2.2) {Activations space:};
            \draw[rounded corners=3pt, gray!40, thin] (-0.15, 0.25) rectangle (3.8,2.5);
            
            \draw[-{Stealth[length=2mm, width=1.5mm]}, thick, langblue] (0,1.6) -- (0.7,1.6);
            \node[font=\footnotesize\sffamily, anchor=west] at (0.85,1.6) {Language};
            
            \draw[-{Stealth[length=2mm, width=1.5mm]}, thick, cultred] (0,1.0) -- (0.7,1.0);
            \node[font=\footnotesize\sffamily, anchor=west] at (0.85,1.0) {Culture};
            
            
        \end{scope}
        
    \end{tikzpicture}
    \caption{Motivation: if culture overlaps with language representations and language identity forms stable directions, then steering with language vectors may improve access to culturally relevant knowledge.}
    \label{fig:vector_types}
\end{figure}
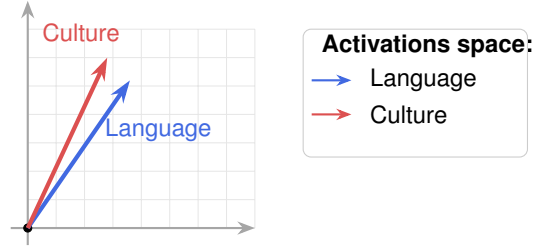

Large language models (LLMs) are increasingly deployed in multilingual settings, but strong multilingual performance does not necessarily imply strong \emph{cultural} competence. Recent work shows that LLMs often underperform on culturally grounded reasoning and everyday cultural knowledge, especially for underrepresented regions and languages, even when they appear linguistically fluent \citep{myung2024blend,romero2024culturalbench}. These concerns have motivated a growing body of research on \emph{cultural awareness} and its evaluation in language models \citep{pawar2024surveyculturalawarenesslanguage}. This challenge is central to SemEval-2026 Task~7 \citep{semeval2026task7}, which evaluates cultural knowledge and reasoning across diverse languages and cultures using BLEnD-style evaluation protocols \citep{myung2024blend}.


In this paper, we describe the \textbf{DFKI-MLT} submission to SemEval-2026 Task~7 \citep{semeval2026task7,semeval-ws-2026-1}. Prior work provides mechanistic evidence that multilingual LLMs encode cultural information in representations that overlap and interact with language-specific components \citep{namazifard2025isolating}, suggesting that intervening on \emph{language-aligned directions} may also modulate culturally relevant behavior. Motivated by this, our system uses \emph{activation steering}: instead of optimizing model parameters through fine-tuning, we modify internal activations at inference time using steering vectors \citep{rimsky-etal-2024-steering}. Concretely, we extract language steering vectors and inject them into the residual stream of multilingual LLMs during generation. We build on evidence that language identity is encoded as a stable direction in activation space \citep{marks2023geometry}, and hypothesize that steering along such directions can improve access to culturally relevant knowledge (Figure~\ref{fig:vector_types}).


Our experiments across multiple multilingual instruction-tuned models, prompts and languages show that activation steering yields \emph{modest} and \emph{heterogeneous} effects on cultural reasoning: at best, we observe improvements of up to \textbf{+1.5\%} absolute accuracy over the unsteered baseline on individual locales, but other configurations degrade performance, and gains do not generalize uniformly across language-region pairs. These results highlight both the appeal of steering as a lightweight inference-time intervention and its current limitations as a stand-alone solution to cultural alignment.

Beyond reporting shared-task performance, we aim to provide a detailed analysis of \emph{when} and \emph{why} using language vectors for activation steering can help cultural reasoning.

\section{Task Background}

SemEval-2026 Task~7 \citep{semeval2026task7} evaluates the \emph{cultural awareness} of language models and NLP systems across languages and regions. The task is based on the manually constructed BLEnD benchmark \citep{myung2024blend}, which is designed specifically for evaluation and therefore does not provide training data. By withholding BLEnD from system training, the shared task aims to assess whether models can generalize to unseen cultural and linguistic contexts rather than memorizing benchmark content.

BLEnD currently covers multiple languages and cultures, and the shared task further expands coverage by adding additional language-culture pairs. Participants may compete in one or more tracks.

\paragraph{Track 1: Short Answer Questions (SAQ).}
In the SAQ track, systems answer short questions in the same language as the input question. The goal is to generate a culturally appropriate response while respecting linguistic and regional variation. Answers are evaluated against human-annotated BLEnD responses.


\paragraph{Track 2: Multiple-Choice Questions (MCQ).}
In the MCQ track, questions are provided in English, and each question includes four answer options representing different cultural perspectives (one option per country/region candidate, subject to the benchmark construction constraints). Systems must select the culturally appropriate option for the target region.

\begin{tcolorbox}[colback=RoyalBlue!5!white, colframe=RoyalBlue!70!white,
                  title=MCQ Example, fonttitle=\bfseries,
                  boxrule=0.5pt, arc=4pt, left=4pt, right=4pt, top=4pt, bottom=4pt]
\small
Question: What sports do men like to watch the most in Ireland?	A.baseball	B.basketball	C.cricket	D.football \\
Gold label: D
\end{tcolorbox}

\paragraph{Our participation.}
We participated in \textbf{both} Track~1 (SAQ) and Track~2 (MCQ). Our submission uses inference-time activation steering with language vectors extracted from multilingual parallel data, without model fine-tuning.

\paragraph{Evaluation metric.}
The official metric is \textbf{accuracy}, with evaluation designed to account for valid response variation. In the SAQ track, a generated answer is considered correct if it matches any acceptable human-annotated response for the same question. In the MCQ track, accuracy is computed based on whether the selected option matches the correct culturally appropriate choice.

\section{System Overview}

Our SemEval-2026 Task 7 submission uses \emph{activation steering} as an inference-time intervention for culturally aware multilingual inference.
Instead of fine-tuning model parameters, we intervene at inference time by adding a steering vector to the residual stream at a selected transformer layer.

The central hypothesis is that language identity is encoded as a direction in activation space \citep{marks2023geometry} and that 
steering along this direction may modulate access to culturally relevant knowledge for a target language-region pair.
We therefore construct language vectors from multilingual sentence representations and inject them during decoding.

The system has three components:
\begin{enumerate}
    \item \textbf{Off-line Language vector extraction} from FLORES-based multilingual data;
    \item \textbf{Inference-time activation steering} with tunable strength $\beta$;
    \item \textbf{Development-time model / layer selection / steering strength} using the SemEval-2026 development phase.
\end{enumerate}

In the final submission, we selected a single steering configuration based on development performance and applied it to both shared-task tracks.
\subsection{Language Vector Extraction}

We compute language vectors from FLORES \citep{nllb2022} sentences by averaging residual-stream activations and taking a difference of means similar to the approach used in AxBench \citep{wu2025axbench}. Let $h^{(l)}(x)$ denote the post-normalization residual-stream activation at layer $l$ for input sentence $x$. For a target language $\ell$, the language vector is defined as:
\begin{equation}
v_{\ell}^{(l)} =
\frac{1}{|D_{\ell}|}\sum_{x \in D_{\ell}} h^{(l)}(x)
-
\frac{1}{|D_{\neg \ell}|}\sum_{x \in D_{\neg \ell}} h^{(l)}(x),
\end{equation}
where $D_{\ell}$ is the set of sentences for the target language and $D_{\neg \ell}$ is the set of sentences from the remaining languages.

\paragraph{Activation extraction details:}
We use the \textbf{post-normalization residual stream} and compute the mean activation over \textbf{all tokens} in each sentence. Sentences are processed one at a time, and no additional prompt template is used during vector extraction (i.e., we feed the original FLORES sentence directly).

\paragraph{FLORES mapping to shared-task language-region pairs:}
BLEnD targets language-region pairs (e.g., \texttt{ar-DZ}, \texttt{es-MX}), while FLORES \citep{nllb2022} provides language/script identifiers. We therefore define a mapping from shared-task pairs to FLORES language codes. For some cases where an exact regional mapping is unavailable in FLORES (e.g., multiple regions sharing the same language variety), we approximate using the closest available language-level FLORES code (e.g., a shared Spanish code for multiple Spanish-speaking regions). We provide the full mapping in Appendix~\ref{app:flores-mapping}.

\paragraph{Data size and preprocessing:}
For each mapped language, we use the first \textbf{1,000} available FLORES dev sentences \citep{nllb2022} to compute the vector. We do not apply additional preprocessing beyond standard tokenization by the model tokenizer. A sample-size convergence study in Appendix~\ref{app:flores-convergence} shows that the resulting DiffMean directions are already highly stable at substantially smaller sample sizes across the models we analyze.

\subsection{Inference-Time Steering}

During inference, we steer the hidden state at a selected transformer layer:
\begin{equation}
\tilde{h}^{(l)} = h^{(l)} + \beta \cdot v_{\ell}^{(l)},
\end{equation}
where $v_{\ell}^{(l)}$ is the language vector for the target language and $\beta$ is a scalar steering strength.

We evaluated a small set of steering strengths $\beta \in \{1, 3, 5\}$ during development and find that \textbf{$\beta=1$} performs best for cultural steering in our setting. This value is used in the final submission.

\subsection{Development-Time Model and Layer Selection}


We perform model and layer selection during the SemEval development phase by evaluating a set of multilingual instruction-tuned LLMs and candidate steering layers. We tested older and newer models in different sizes that have proven to perform well in multilingual settings, including Qwen2.5-72B-Instruct and Qwen2.5-7B-Instruct \citep{qwen2.5}, Aya Expanse 8B and Aya Expanse 32B \citep{dang2024ayaexpansecombiningresearch}, and Qwen3-8B and Qwen3-32B \citep{qwen3technicalreport}.

Based on development performance, we select \textbf{Qwen2.5-72B-Instruct} with steering applied at \textbf{Layer~26} for the final shared-task submission.

\section{Experimental Setup}




\subsection{Decoding and Inference}

We use greedy decoding (\texttt{temperature}=0) for both tracks to minimize confounding factors when evaluating activation steering. Since our method intervenes directly on internal representations, stochastic decoding (e.g., sampling with nonzero temperature) would introduce additional variance that can obscure whether performance changes are caused by the intervention or by decoding randomness. Deterministic decoding therefore allows a clearer attribution of gains or degradations to the steering configuration (layer and $\beta$), and improves reproducibility across layer sweeps and prompt comparisons.

\paragraph{Track 2 (MCQ).}
For each question, we prompt the model to choose one option from \texttt{A/B/C/D}. We score the four answer letters using their \textbf{output log-probabilities} and select the option with the highest log-probability. We generate at most \textbf{1 token}.

\paragraph{Track 1 (SAQ).}

We generate up to \textbf{32 tokens} to balance completeness and evaluation stability. Although SAQ targets concise answers, the required length varies across languages due to tokenization and morphology (e.g., multi-word expressions), and overly small limits risk truncating otherwise correct answers. At the same time, longer generations increase the chance of irrelevant continuations that can hurt near-exact matching. To reduce formatting artifacts, we apply a lightweight normalization procedure to the generated text (Normalization details in Appendix~\ref{sec:saq-postprocessing}).




\subsection{Prompting Strategy}

We evaluate two prompt formulations for both tracks during analysis: a \textbf{generic prompt} and a \textbf{cultural prompt}. The official shared-task submission uses the \textbf{cultural prompt}.

\paragraph{Generic prompt.}
The generic prompt instructs the model to answer the question (or select one MCQ option) without explicitly mentioning the target region or language in the instruction text.

\begin{tcolorbox}[colback=RoyalBlue!5!white, colframe=RoyalBlue!70!white,
                  title=Generic prompt Template, fonttitle=\bfseries,
                  boxrule=0.5pt, arc=4pt, left=4pt, right=4pt, top=4pt, bottom=4pt]
\small
Select exactly one option: A, B, C, or D. \\ 
Question: \{question\} \\
A. \{option\_a\} \\
B. \{option\_b\} \\
C. \{option\_c\} \\
D. \{option\_d\} \\
Answer (A/B/C/D):
\end{tcolorbox}

\paragraph{Cultural prompt (official submission).}
The cultural prompt explicitly conditions the model on the target region and language (e.g., ``for someone living in [region]'' and ``respond in [language]''). For SAQ, it additionally instructs the model to produce a concise answer without explanation.

\begin{tcolorbox}[colback=RoyalBlue!5!white, colframe=RoyalBlue!70!white,
                  title=Cultural prompt Template, fonttitle=\bfseries,
                  boxrule=0.5pt, arc=4pt, left=4pt, right=4pt, top=4pt, bottom=4pt]
\small
You are answering a multiple-choice question for someone living in \{Region\}. 
Respond strictly in \{Language\} and select exactly one option: A, B, C, or D. \\ 
Question: \{question\} \\
A. \{option\_a\} \\
B. \{option\_b\} \\
C. \{option\_c\} \\
D. \{option\_d\} \\
Answer (A/B/C/D):
\end{tcolorbox}



\subsection{Hyperparameters}

We select the steering strength from $\beta \in \{1,3,5\}$ on the SemEval-2026 development phase and use $\beta=1$ in the final submission. We run layer sweeps to locate the best depth for steering. The steering layer (Layer~26) and backbone model (Qwen2.5-72B-Instruct) are also chosen based on development performance for the official results.



\section{Results and Analysis}

\begin{table}[t]
\centering
\small
\begin{tabular}{lccc}
\toprule
\textbf{Track} & \textbf{Metric} & \textbf{Score} & \textbf{Rank} \\
\midrule
Track 1 (SAQ) & Acc. & N/A & -- / 10 \\
Track 2 (MCQ) & Acc. & 86.96 & 7 / 17 \\
\bottomrule
\end{tabular}
\caption{Official SemEval-2026 Task~7 results for our submission. The official submission used the \textbf{cultural prompt}. Our SAQ submission was not evaluated due to a corrupted/incorrect file and therefore has no official score.}
\label{tab:official-results}
\end{table}

\begin{table}[t]
\centering
\small
\begin{tabular}{lccc c}
\toprule
\textbf{Locale} & \textbf{Ours (\%)} & \textbf{Our rank} & \textbf{Best (\%)} & \textbf{Gap} \\
\midrule
\texttt{es-EC} & 97.54 & 7 & 98.67 & $-1.13$ \\
\texttt{en-GB} & 96.12 & 6 & 99.17 & $-3.05$ \\
\texttt{es-MX} & 94.94 & 4 & 99.32 & $-4.38$ \\
\texttt{ar-EG} & 94.84 & 2 & 91.03 & $+3.81$ \\
\texttt{bg-BG} & 94.60 & 8 & 99.54 & $-4.94$ \\
\bottomrule
\end{tabular}
\caption{Top-5 language--region pairs (Track~2 MCQ) by our official accuracy. ``Best'' denotes the top-ranked system on the leaderboard (overall winner). Positive gap indicates our score exceeds the winner's per-locale score in the excerpt.}
\label{tab:mcq-top5}
\end{table}

\subsection{Official Shared-Task Results}

Due to an incorrect/corrupted submission file, our Track~1 (SAQ) submission was not successfully evaluated by the organizers and therefore has no official score. We therefore report official leaderboard results only for Track~2 (MCQ).

Using the cultural prompt and activation steering, our Track~2 system achieved \textbf{86.96\%} overall accuracy and ranked \textbf{7th} out of \textbf{17} teams (Table~\ref{tab:official-results}). The best-performing system on the leaderboard achieved \textbf{96.78\%}, leaving a gap of \textbf{9.82} percentage points to our submission. 

Table~\ref{tab:mcq-top5} lists five language-region pairs where our system performs best at the MCQ track. For each locale, we report our accuracy and our locale-specific rank (based on the official leaderboard). For \texttt{ar-EG}, we obtain \textbf{94.84\%} while the overall winner system reports \textbf{91.03\%}, meaning we outperform the winning system by \textbf{3.81} percentage points on this locale. In contrast, for \texttt{bg-BG}, we reach \textbf{94.60\%} while the winner achieves \textbf{99.54\%}, leaving a \textbf{4.94} percentage-point deficit. This heterogeneity aligns with our post-hoc analyses, which indicate that both steering and prompting effects are highly locale-dependent. 


\subsection{Post-hoc Analysis}
\label{sec:posthoc-summary}

\begin{figure}[t]
    \centering
    \includegraphics[width=0.9\linewidth]{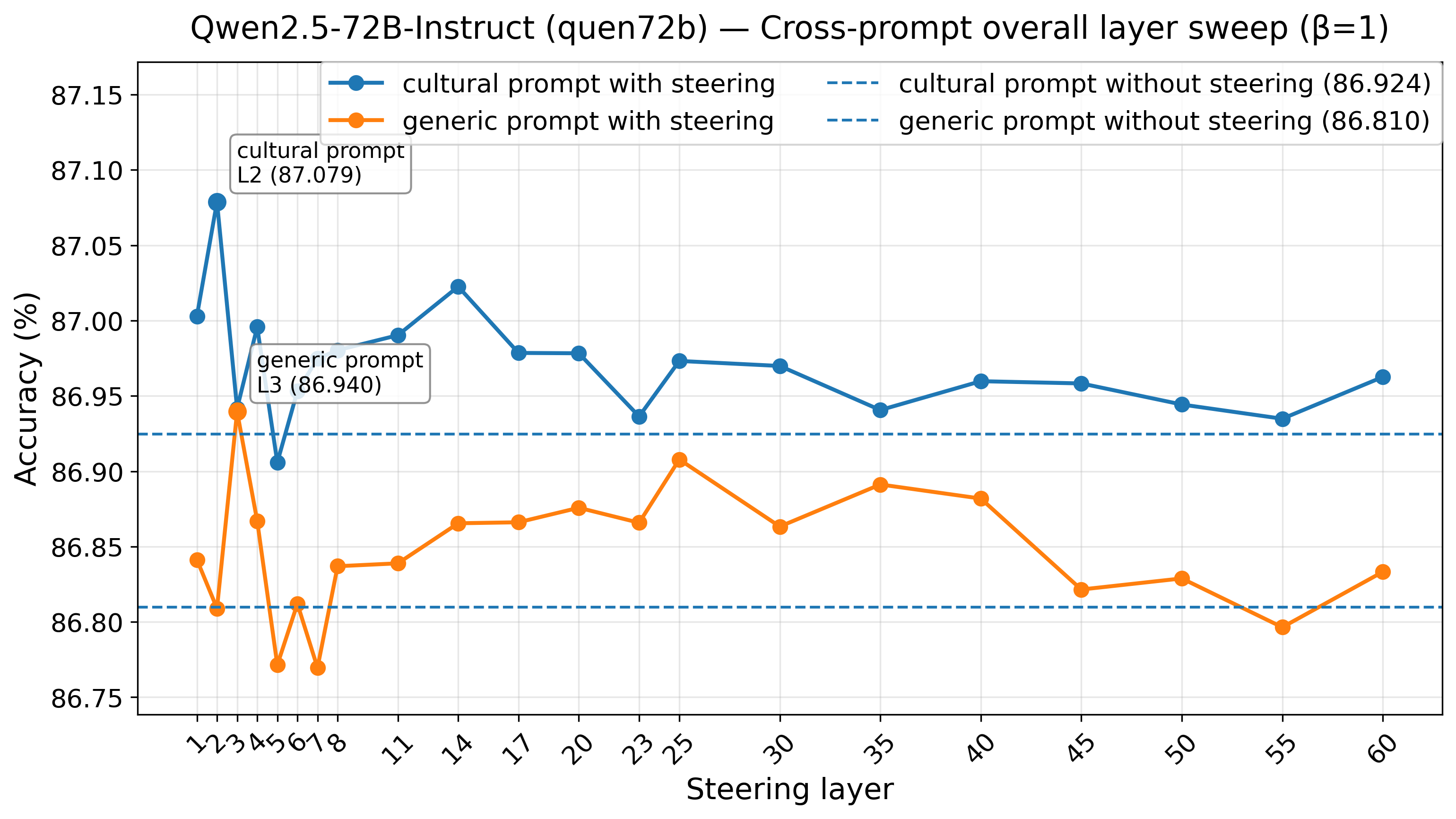}
    \includegraphics[width=0.9\linewidth]{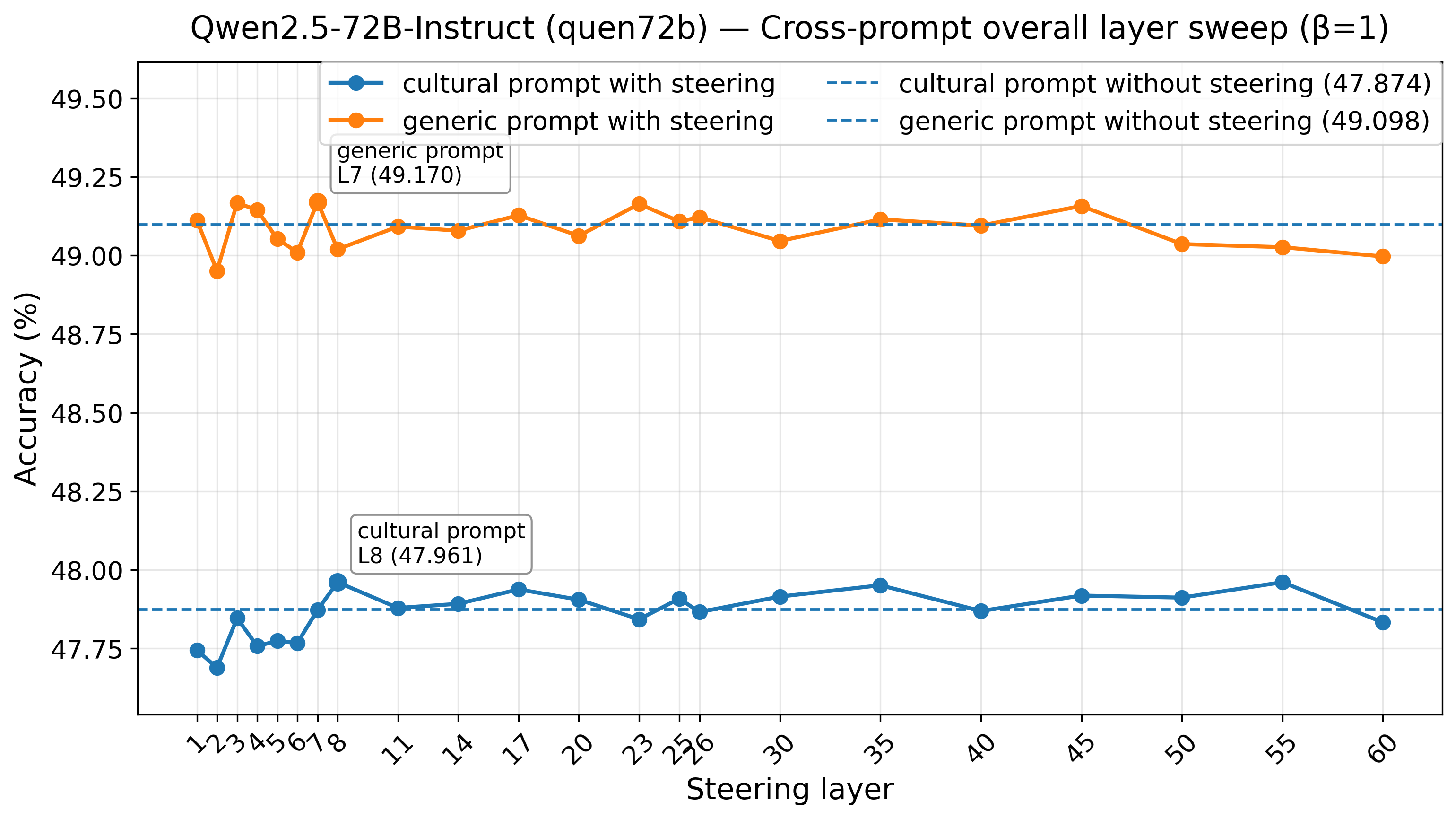}
    
    \caption{Post-hoc cross-prompt layer sweeps for Qwen2.5-72B-Instruct with $\beta=1$ on \textbf{MCQ} (top) and \textbf{SAQ} (bottom). The official submission uses the \textbf{cultural prompt}.}
    \label{fig:crossprompt-layer-sweep-both}

\end{figure}

\begin{figure}[t]
    \centering
    \includegraphics[width=0.9\linewidth]{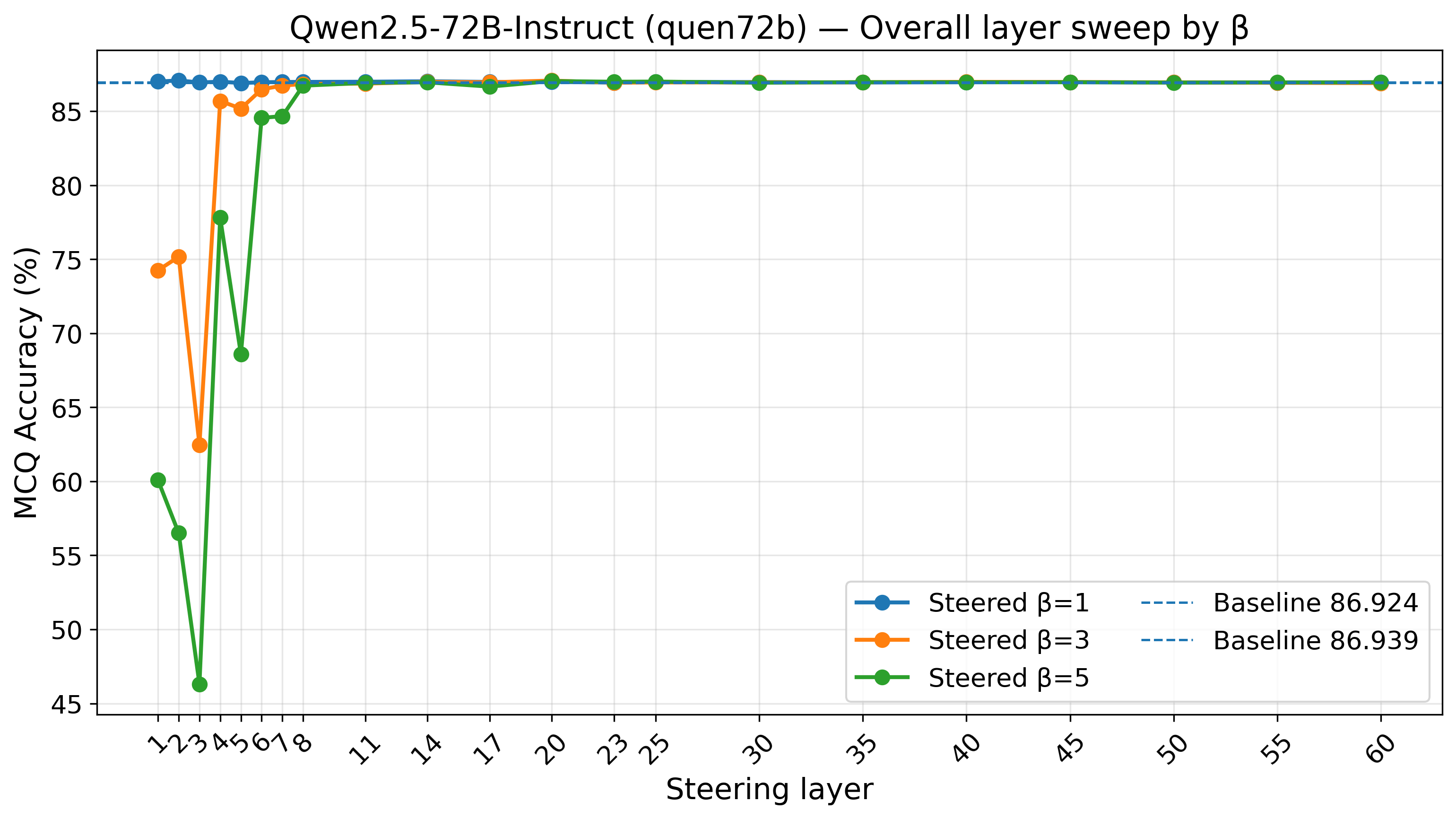}
    \caption{Post-hoc overall MCQ layer sweeps for Qwen2.5-72B-Instruct under different steering strengths ($\beta \in \{1,3,5\}$).}
    \label{fig:qwen72b-multibeta-layer-sweep}
\end{figure}

\begin{figure}[t]
    \centering
    \includegraphics[width=0.9\linewidth]{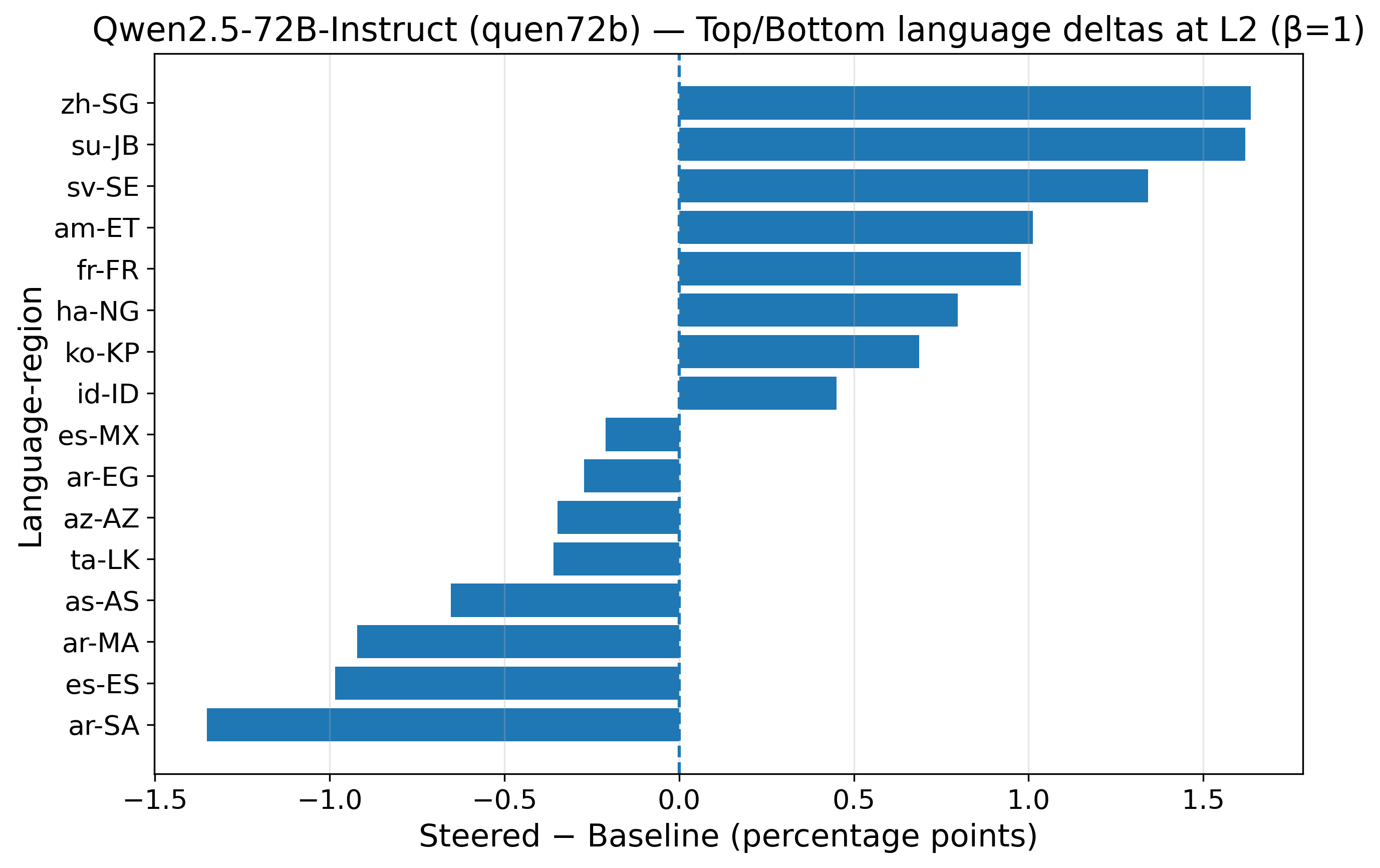}
    \caption{Top and bottom per-language MCQ accuracy changes (steered minus baseline, percentage points) using the cultural prompt.}
    \label{fig:qwen72b-topbottom-l2}
\end{figure}

To characterize system behavior beyond the single locked submission configuration, we ran post-hoc analyses on both MCQ and SAQ evaluation data across multiple models (Qwen2.5-72B/7B, Aya Expanse 8B/32B, Qwen3 8B/32B) using the same evaluation metrics provided by the SemEval-2026 organizers for each track.
We observe: 

(i) \textbf{strong layer sensitivity}: steering gains concentrate in a subset of layers while some layers degrade performance (e.g., the MCQ/SAQ cross-prompt layer sweeps in Figure~\ref{fig:crossprompt-layer-sweep-both}); notably, for Qwen2.5-72B the best steering layer differs by both task and prompt (MCQ peaks at Layer~2 vs.\ 3, and SAQ peaks at Layer~8 vs.\ 7 for cultural vs.\ generic), illustrating that a single global layer choice is a compromise.
Notably, the best post-hoc layer differs from the layer selected for the official submission due to differences in evaluation split.

(ii) \textbf{$\beta$ sensitivity}: larger steering strengths are more prone to early-layer instability, whereas smaller strengths are generally more robust; in practice we found $\beta=1$ to be the most reliable setting for Qwen2.5 models (Figure~\ref{fig:qwen72b-multibeta-layer-sweep}), while some Qwen3/Aya configurations tolerate stronger steering in post-hoc sweeps (Appendices~\ref{app:posthoc-plots_mcq} and~\ref{app:posthoc-plots-saq}).

(iii) \textbf{prompt-task interaction}: cultural prompting tends to be stronger for MCQ, where it conditions choice probabilities without changing output format, whereas the generic prompt is often better for SAQ across several models, likely because SAQ scoring depends on matching short surface forms and culturally conditioned prompts can induce verbose or stylistically marked responses (Figure~\ref{fig:crossprompt-layer-sweep-both}). For example, for the SAQ item \emph{``What is a popular snack at an amusement park in Azerbaijan?''}, the generic prompt yields a short candidate (\emph{``Somsa/Samsa''}), while the cultural prompt produces a longer explanatory response (e.g., \emph{``A popular snack at an amusement park in Azerbaijan is pakhlava, a sweet pastry...''}), which is more likely to fail evaluation even when broadly plausible. This aligns with findings that cultural prompting can be beneficial but is not uniformly effective across settings \citep{tao2024cultural}.

(iv) \textbf{model- and locale-dependent effects}: steering impacts vary substantially across language-region pairs (Figure~\ref{fig:qwen72b-topbottom-l2}), with some locales showing large gains and others degradations, and these patterns are not uniform across models, motivating model- and locale-aware steering policies in future work.

(v) \textbf{model- and $\beta$ effects}: 
We do not observe a simple monotonic relationship between model parameter count or depth and the optimal steering strength $\beta$ in our post-hoc sweeps. The preferred $\beta$ appears model- and setting-dependent: across our evaluated models, $\beta=1$ is the safest default, while a few Qwen3/Aya configurations tolerate stronger steering in localized layers (Appendices~\ref{app:posthoc-plots_mcq} and~\ref{app:posthoc-plots-saq}). We therefore caution against treating $\beta$ as a function of scale alone, and recommend re-tuning it per model and prompt.

\begin{figure}[t]
    \centering
    \includegraphics[width=0.95\linewidth]{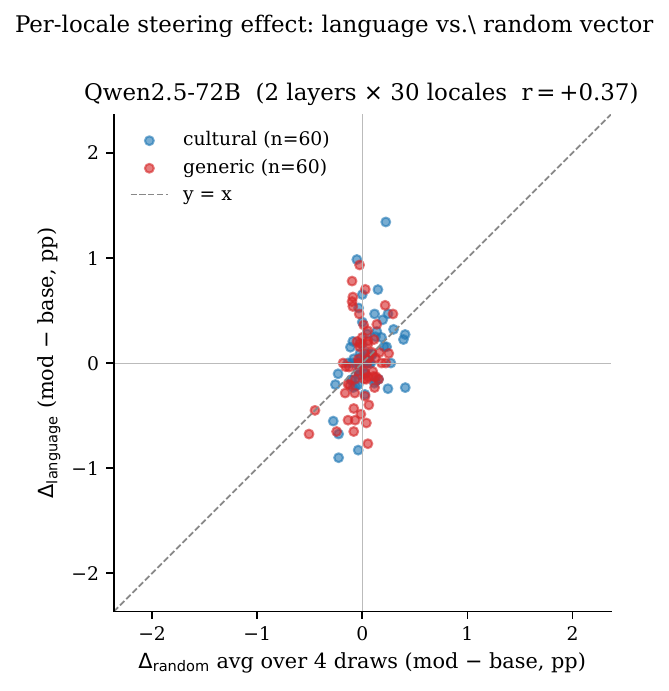}
    \caption{Per-locale steering effect for Qwen2.5-72B: $\Delta_{\mathrm{random}}$ averaged over four Gaussian draws (x-axis) vs.\ $\Delta_{\mathrm{language}}$ (y-axis), using the two dev-selected layers and both prompts. Each point is a (layer, prompt, locale) cell; $n{=}60$ per prompt. Random-vector effects concentrate near zero, while language-vector effects span a wider range and include negative outliers.}
    \label{fig:rand-vs-lang-avg-scatter}
\end{figure}

(vi) \textbf{random vs.\ language vector effects}: To check whether language-vector effects are distinguishable from generic activation perturbations, we compare them against L2-normalized Gaussian random vectors at the same layers with the same $\beta=1$ intervention (Appendix~F). For Qwen2.5-72B, random-vector effects remain concentrated near zero after averaging over four draws, while language-vector effects are somewhat more dispersed and include negative outliers (Figure~\ref{fig:rand-vs-lang-avg-scatter}). This suggests that random perturbations do not fully explain the language-vector effects, but the effects are also not reliably beneficial.

\section{Discussion}

Our post-hoc analyses indicate that activation steering for cultural MCQ/SAQ reasoning yields modest and highly context-dependent improvements rather than uniform gains. First, the steering effect is strongly layer-sensitive, with improvements concentrated in a subset of layers and other layers degrading performance. Second, per-config means stay under 0.5 pp on either track because most layers are neutral, and gains on one track do not predict gains on the other, so a single global steering layer (e.g., the Layer 26 used for the official submission) cannot be optimal for every (locale, track) pair. Third, prompt design interacts with steering in non-trivial ways: the cultural prompt used for the official submission and a simpler generic prompt produce different optimal steering layers and different per-language gains. 

These findings indicate that prompt design and activation steering should be treated as a jointly optimized inference-time adaptation problem rather than independent components.

\section{Limitations and Future Work}
\label{sec:limitations}

\begin{itemize}

\item \textbf{Official evaluation coverage.} Our Track~1 (SAQ) submission was not officially evaluated because of a corrupted file. All SAQ results are therefore \emph{post-hoc} offline re-evaluations and not comparable to the official leaderboard. Future submissions should include stricter package validation before upload.

\item \textbf{Scope of empirical comparison.} We analyze sensitivity to layer, $\beta$, prompt, model, and locale, but do not exhaustively compare against stronger prompt-only baselines, fine-tuning, or alternative steering methods such as CAA, ReFT, or SAE-based steering. Future work should benchmark DiffMean steering against these adaptation methods under matched compute and evaluation settings.

\item \textbf{Language-derived vectors and cultural conflation.} Our vectors are derived from FLORES language-level data rather than culturally annotated or task-specific data. This conflates language identity with culture: several language--region pairs share the same FLORES code, so within-language regional variation is not captured. Future work should compare FLORES-based vectors with culture-specific and task-specific steering directions.

\item \textbf{Single global steering configuration.} Our official submission uses one global $(\beta,\text{layer})$ pair, although post-hoc analyses show that locally optimal settings vary across models, prompts, layers, and locales. Future work should explore adaptive per-language, per-locale, or per-prompt steering policies.

\end{itemize}



\newpage
\section*{Acknowledgments}
This research was supported by the German Federal Ministry of Research, Technology and Space (BMFTR) as part of the
project TRAILS (01IW24005).

\bibliography{custom}

\appendix
\section*{Appendix}

\section{FLORES Mapping for Shared-Task language-region Pairs}
\label{app:flores-mapping}

We map BLEnD language-region pairs to FLORES language/script identifiers to compute language vectors. In cases where FLORES does not provide a region-specific variety, we use the closest available language-level approximation (e.g., a shared Spanish FLORES code for multiple Spanish-speaking regions). Table~\ref{tab:flores-mapping} provides the complete mapping from BLEnD language-region pairs to FLORES language/script identifiers used to compute language vectors.

\begin{table*}[t]
\centering
\tiny
\begin{tabular}{ll ll ll ll}
\toprule
\textbf{BLEnD locale} & \textbf{FLORES code} &
\textbf{BLEnD locale} & \textbf{FLORES code} &
\textbf{BLEnD locale} & \textbf{FLORES code} &
\textbf{BLEnD locale} & \textbf{FLORES code} \\
\midrule
\texttt{am-ET} & \texttt{amh\_Ethi} &
\texttt{ar-DZ} & \texttt{kab\_Latn} &
\texttt{ar-EG} & \texttt{arz\_Arab} &
\texttt{ar-MA} & \texttt{ary\_Arab} \\
\texttt{ar-SA} & \texttt{ars\_Arab} &
\texttt{as-AS} & \texttt{asm\_Beng} &
\texttt{az-AZ} & \texttt{azj\_Latn} &
\texttt{bg-BG} & \texttt{bul\_Cyrl} \\
\texttt{el-GR} & \texttt{ell\_Grek} &
\texttt{en-AU} & \texttt{eng\_Latn} &
\texttt{en-GB} & \texttt{eng\_Latn} &
\texttt{en-US} & \texttt{eng\_Latn} \\
\texttt{es-EC} & \texttt{spa\_Latn} &
\texttt{es-ES} & \texttt{spa\_Latn} &
\texttt{es-MX} & \texttt{spa\_Latn} &
\texttt{eu-ES} & \texttt{eus\_Latn} \\
\texttt{eu-PV} & \texttt{eus\_Latn} &
\texttt{fa-IR} & \texttt{pes\_Arab} &
\texttt{fr-FR} & \texttt{fra\_Latn} &
\texttt{ga-IE} & \texttt{gle\_Latn} \\
\texttt{ha-NG} & \texttt{hau\_Latn} &
\texttt{id-ID} & \texttt{ind\_Latn} &
\texttt{ja-JP} & \texttt{jpn\_Jpan} &
\texttt{ko-KP} & \texttt{kor\_Hang} \\
\texttt{ko-KR} & \texttt{kor\_Hang} &
\texttt{ms-SG} & \texttt{zsm\_Latn} &
\texttt{su-JB} & \texttt{jav\_Latn} &
\texttt{sv-SE} & \texttt{swe\_Latn} \\
\texttt{ta-SG} & \texttt{tam\_Taml} &
\texttt{tl-PH} & \texttt{tgl\_Latn} &
\texttt{zh-CN} & \texttt{zho\_Hans} &
\texttt{zh-TW} & \texttt{zho\_Hant} \\
\texttt{zh-SG} & \texttt{zsm\_Latn} &
\texttt{en-AS} & \texttt{asm\_Beng} &
\texttt{en-AZ} & \texttt{azj\_Latn} &
\texttt{en-BG} & \texttt{bul\_Cyrl} \\
\texttt{en-CN} & \texttt{zho\_Hans} &
\texttt{en-DZ} & \texttt{kab\_Latn} &
\texttt{en-EG} & \texttt{arb\_Latn} &
\texttt{en-ES} & \texttt{spa\_Latn} \\
\texttt{en-ET} & \texttt{amh\_Ethi} &
\texttt{en-FR} & \texttt{fra\_Latn} &
\texttt{en-GR} & \texttt{ell\_Grek} &
\texttt{en-ID} & \texttt{ind\_Latn} \\
\texttt{en-IE} & \texttt{gle\_Latn} &
\texttt{en-IR} & \texttt{pes\_Arab} &
\texttt{en-JP} & \texttt{jpn\_Jpan} &
\texttt{en-KP} & \texttt{kor\_Hang} \\
\texttt{en-KR} & \texttt{kor\_Hang} &
\texttt{en-LK} & \texttt{sin\_Sinh} &
\texttt{en-MA} & \texttt{arb\_Latn} &
\texttt{en-MX} & \texttt{spa\_Latn} \\
\texttt{en-NG} & \texttt{hau\_Latn} &
\texttt{en-PH} & \texttt{tgl\_Latn} &
\texttt{en-PV} & \texttt{eus\_Latn} &
\texttt{en-SA} & \texttt{ars\_Arab} \\
\texttt{en-SE} & \texttt{swe\_Latn} &
\texttt{en-SG} & \texttt{tam\_Taml} &
\texttt{en-TW} & \texttt{zho\_Hant} &
\texttt{en-EC} & \texttt{spa\_Latn} \\
\texttt{en-JB} & \texttt{jav\_Latn} &
\texttt{ta-LK} & \texttt{sin\_Sinh} &
 &  &  &  \\
\bottomrule
\end{tabular}
\caption{Complete mapping from BLEnD language--region pairs to FLORES language/script identifiers used for language vector computation. Some mappings are approximations when an exact region-specific FLORES variety is unavailable.}
\label{tab:flores-mapping}
\end{table*}

\section{FLORES Sample-Size Convergence Study}
\label{app:flores-convergence}

We test whether the DiffMean language vectors used in \S3.1 are sensitive to the number of FLORES sentences used for extraction. For each of the six post-hoc models (Qwen2.5-72/7B-Instruct, Qwen3-32/8B, and Aya-Expanse-32/8B) we re-estimate vectors for the same 28 FLORES languages at $N \in \{100,200,\ldots,1000\}$. For each language and layer, we compute $v_N^{(\ell,l)}$ from the first $N$ FLORES dev sentences, using the same post-normalization residual-stream activations and token averaging as in \S3.1. Since each $N$ is a strict prefix of the $N{=}1000$ set, differences from $v_{1000}$ isolate the effect of adding more sentences rather than changing the sample.

\begin{figure*}[t]
    \centering
    \includegraphics[width=0.97\linewidth]{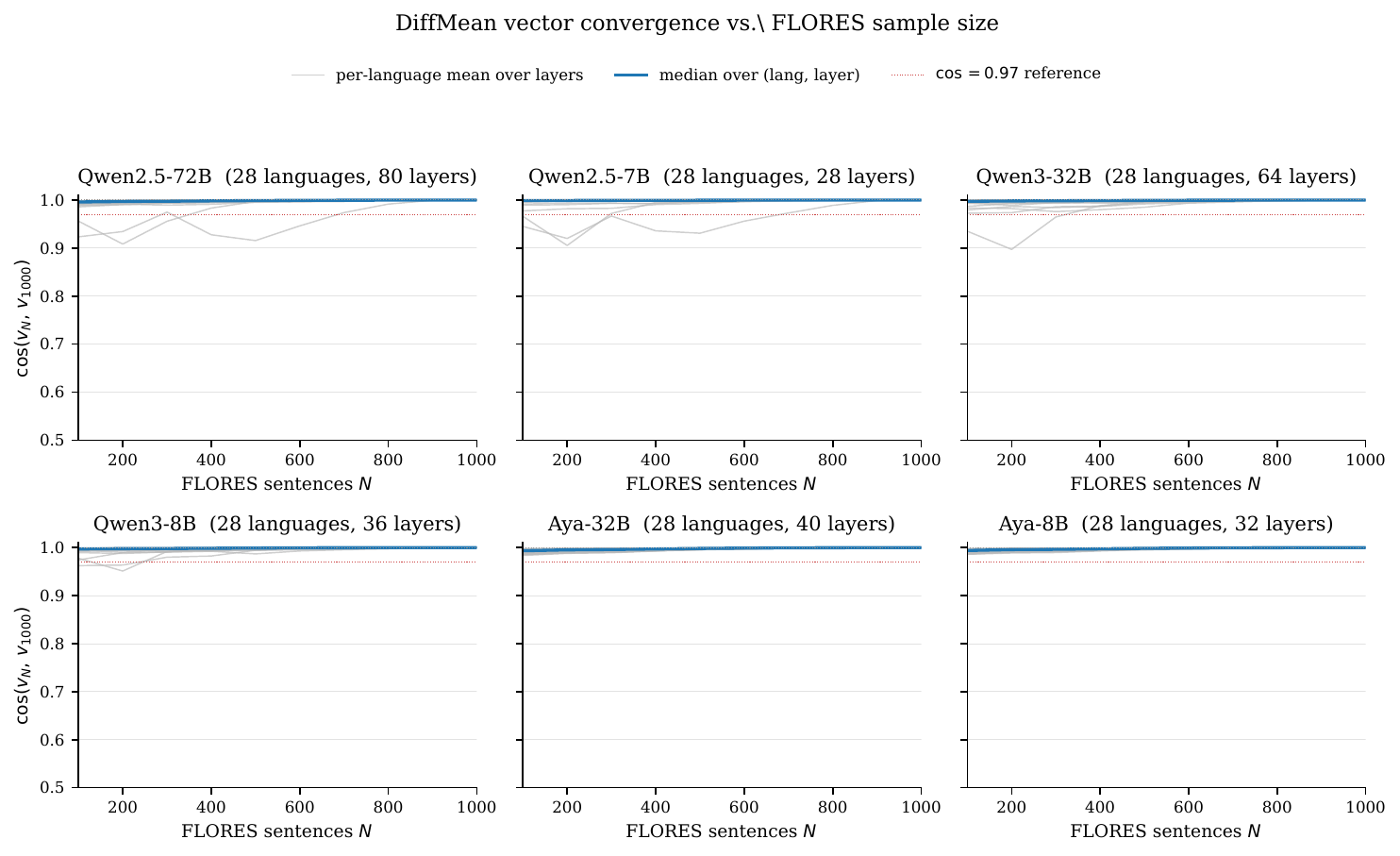}
    \caption{DiffMean vector convergence vs.\ FLORES sample size: $\cos(v_N, v_{1000})$ as a function of $N$ for all six post-hoc models over 28 FLORES languages. Faint grey curves are individual languages, averaged over layers; the bold curve is the joint median over language--layer pairs, with the shaded 25--75\% IQR. The dotted reference line marks $\cos=0.97$, and the $y$-axis is zoomed to show sub-1 variation.}
    \label{fig:flores-convergence-grid}
\end{figure*}

We measure convergence with cosine similarity,
$\cos(v_N^{(\ell,l)}, v_{1000}^{(\ell,l)})$, for every language--layer--$N$ cell. Figure~\ref{fig:flores-convergence-grid} summarizes the results with per-language curves, averaged over layers, and a joint median with 25--75\% IQR over all language--layer cells.

Across all six models, the joint median is already at least $0.99$ at $N{=}100$ and reaches at least $0.999$ by $N{=}500$; the IQR is essentially collapsed near $1.0$ from about $N{=}300$ onward. Low-$N$ outliers are model-dependent: the worst per-language layer means at $N{=}100$ range from about $0.96$ for Aya models to about $0.86$ for Qwen2.5-7B. Qwen2.5-72B does not show worse stability than smaller models, suggesting that greater depth does not require more FLORES data within the tested $[100,1000]$ range.

These results indicate that using $1{,}000$ FLORES sentences is conservative for the six models studied here. Since the downstream intervention uses the unit-norm direction $v$ with $\beta=1$, a cosine similarity of $0.99$ corresponds to a steering-direction change below $\arccos(0.99) \approx 8^\circ$, smaller than the layer-to-layer variation observed in our steering sweeps. We therefore do not expect FLORES sample size to be a major source of instability in our reported results.

\section{Post-processing for SAQ}
\label{sec:saq-postprocessing}

For SAQ evaluation, we normalize model outputs using simple string cleanup heuristics:
\begin{itemize}
    \item truncate at \texttt{<|end\_of\_text|>} if present;
    \item keep only the first line;
    \item keep text before the first period;
    \item collapse repeated whitespace;
    \item remove quotation marks.
\end{itemize}

This post-processing is applied before matching generated answers to the set of human-annotated acceptable responses.

\clearpage


\begin{figure}[t]
\section{Post-hoc MCQ Analysis Plots}
\label{app:posthoc-plots_mcq}
This appendix provides additional post-hoc analysis plots for all tested models.
\subsection{Qwen2.5-72B-Instruct}
    \centering
    \includegraphics[width=0.9\linewidth]{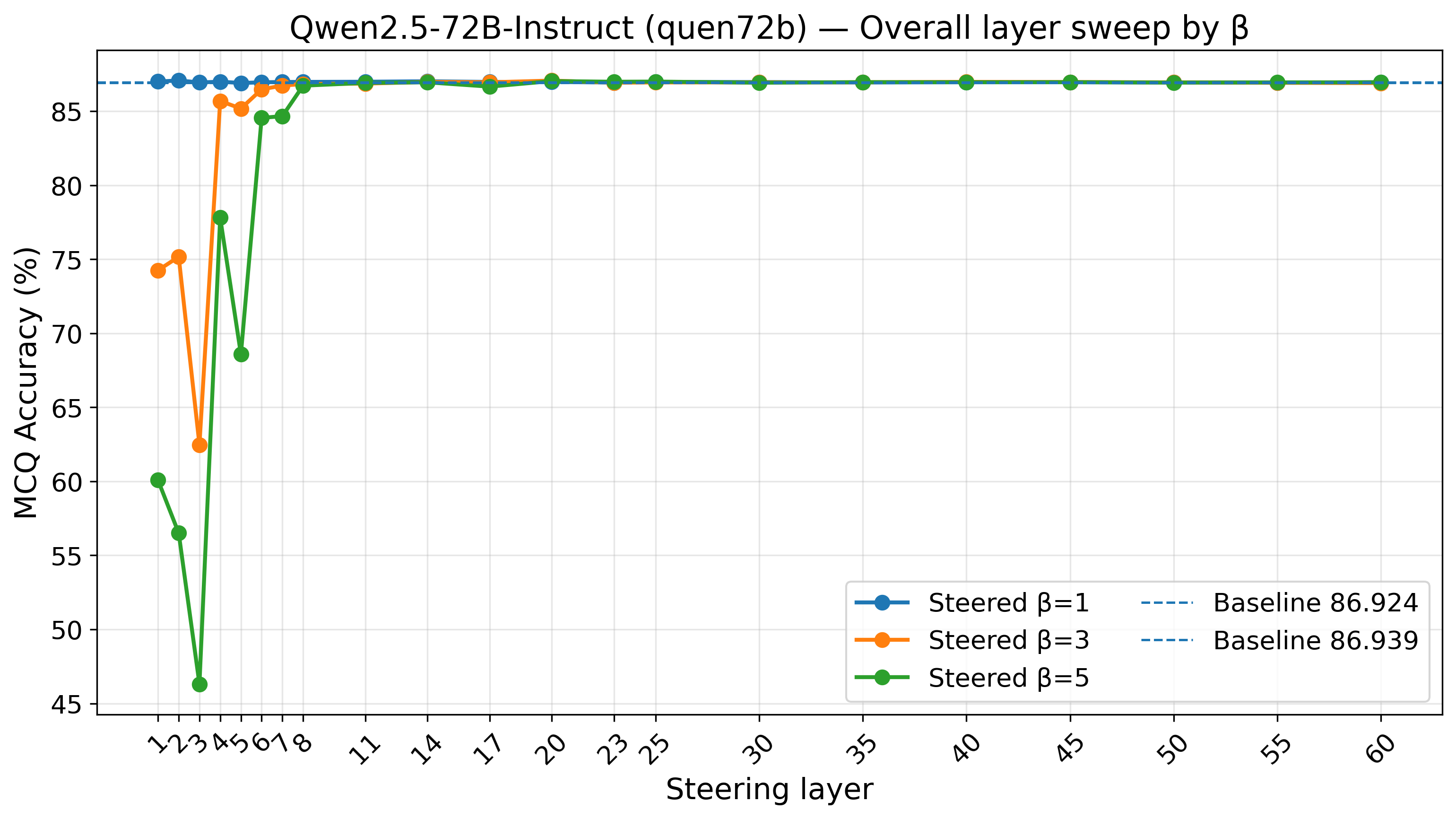}
    \caption{Post-hoc overall MCQ layer sweeps for Qwen2.5-72B-Instruct under different steering strengths ($\beta \in \{1,3,5\}$). Large steering strengths can substantially degrade performance in early layers, while $\beta=1$ remains stable and yields the best overall trade-off in our experiments.}
\end{figure}


\begin{figure}[t]
    \centering
    \includegraphics[width=0.9\linewidth]{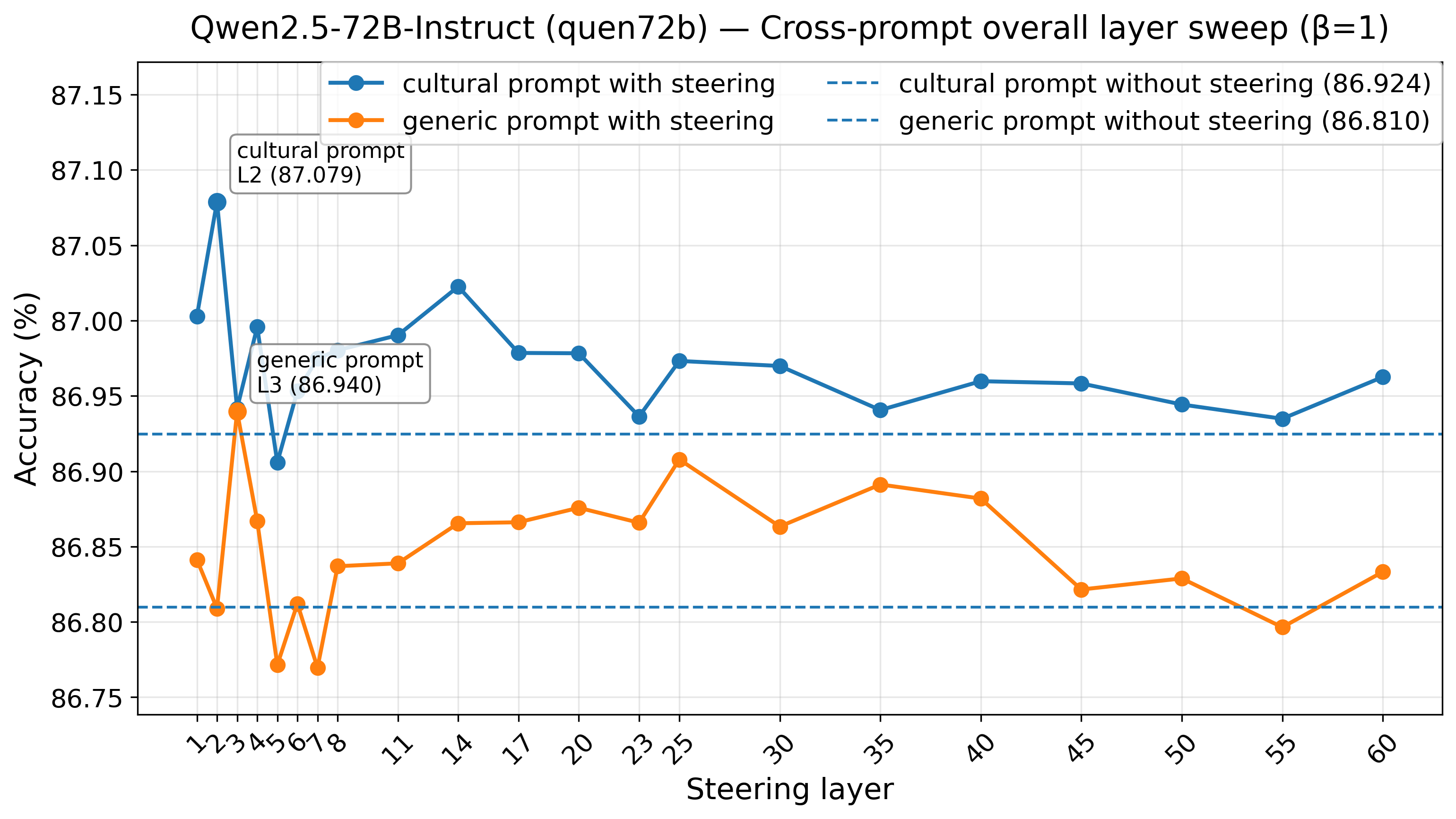}
    \caption{Post-hoc cross-prompt MCQ layer sweep for Qwen2.5-72B-Instruct with $\beta=1$. The official submission uses the \textbf{cultural prompt}. Prompt choice affects both baseline accuracy and the optimal steering layer (here, Layer~2 for the cultural prompt and Layer~3 for the generic prompt).}
\end{figure}


\begin{figure}[t]
    \centering
    \includegraphics[width=0.9\linewidth]{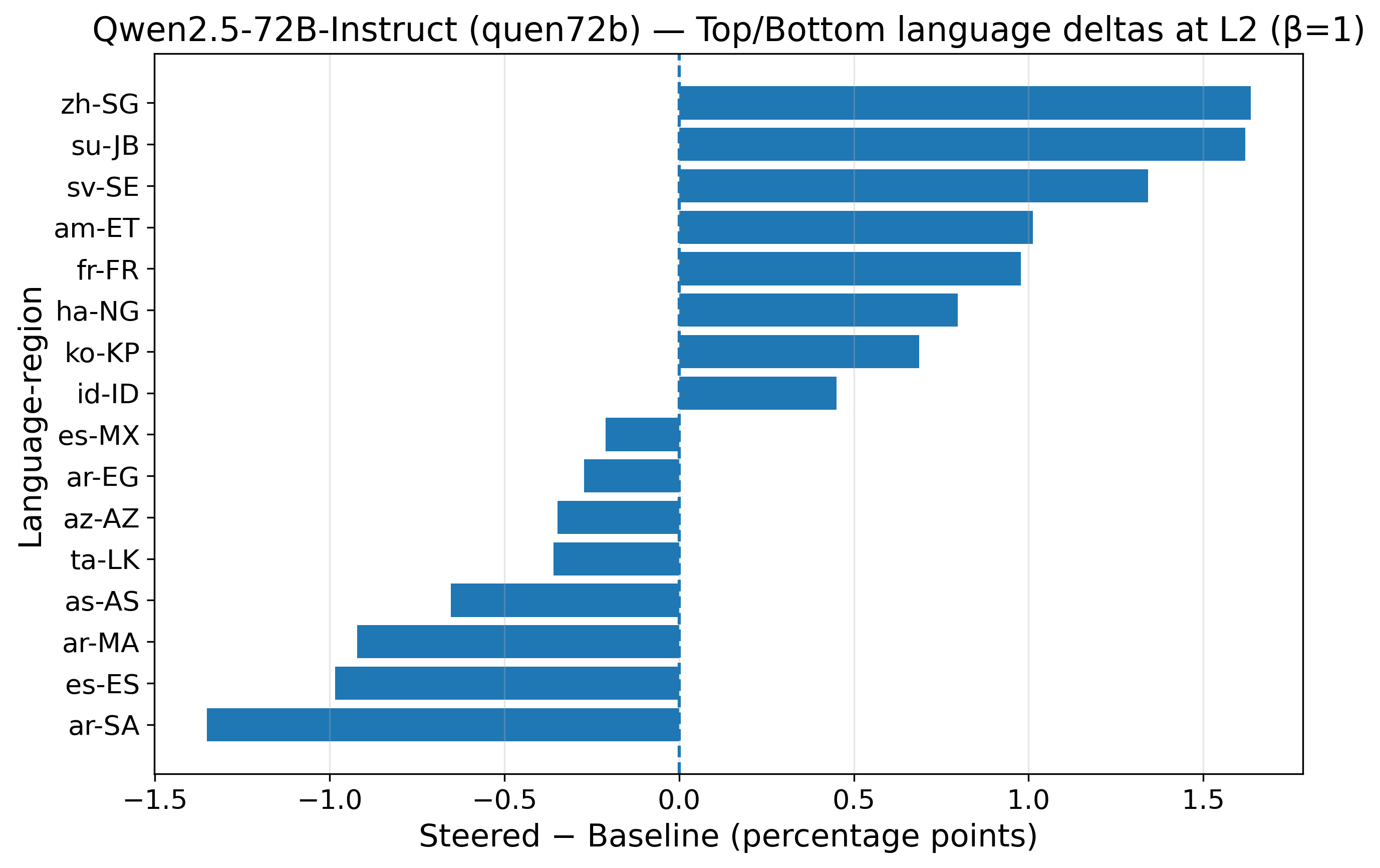}
    \caption{Top and bottom per-language MCQ accuracy changes (steered minus baseline, percentage points) for Qwen2.5-72B-Instruct at Layer~2 with $\beta=1$ using the cultural prompt. Steering produces substantially different effects across language-region pairs, including both strong gains and degradations.}
\end{figure}



\newpage

\begin{figure}[t]
\subsection{Qwen2.5-7B-Instruct}

    \centering
    \includegraphics[width=0.9\linewidth]{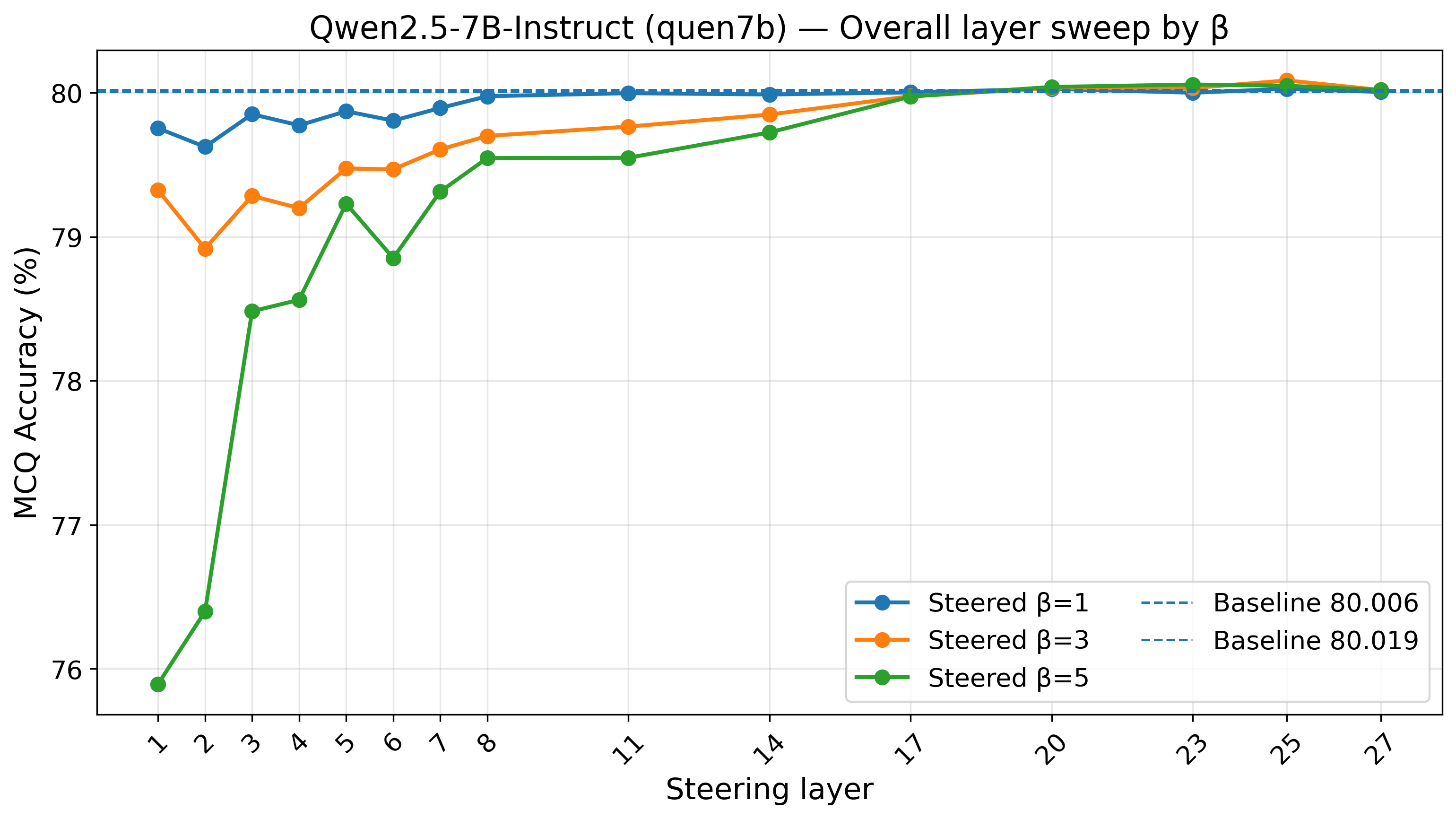}
    \caption{Post-hoc overall MCQ layer sweeps for Qwen2.5-7B-Instruct under different steering strengths ($\beta \in \{1,3,5\}$). Large steering strengths can substantially degrade performance in early layers, while $\beta=1$ remains stable and yields the best overall trade-off in our experiments.}
\end{figure}


\begin{figure}[t]
    \centering
    \includegraphics[width=0.9\linewidth]{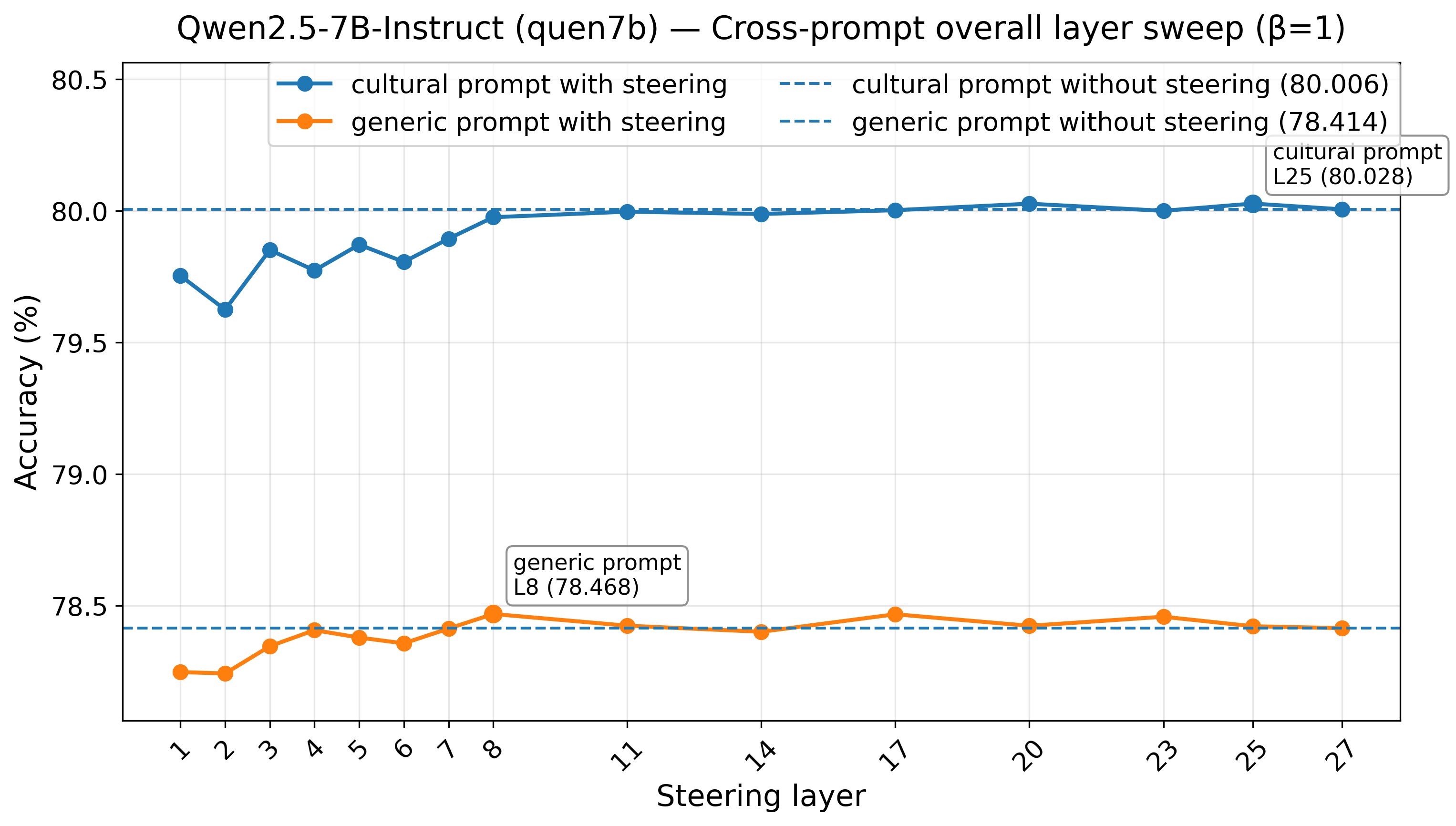}
    \caption{Post-hoc cross-prompt MCQ layer sweep for Qwen2.5-7B-Instruct with $\beta=1$. The official submission uses the \textbf{cultural prompt}. Prompt choice affects both baseline accuracy and the optimal steering layer (here, Layer~25 for the cultural prompt and Layer~8 for the generic prompt).}
\end{figure}


\begin{figure}[t]
    \centering
    \includegraphics[width=0.9\linewidth]{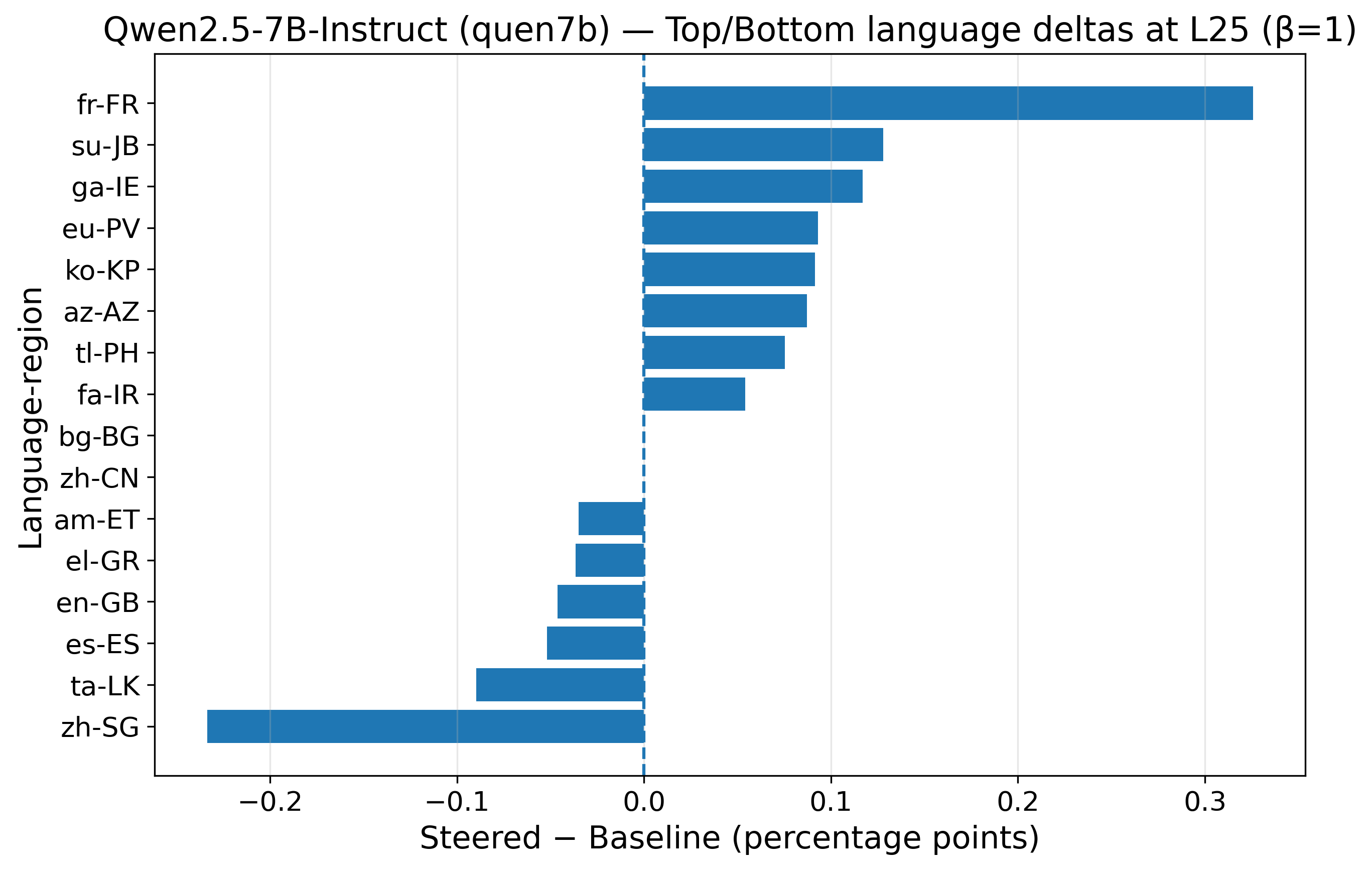}
    \caption{Top and bottom per-language MCQ accuracy changes (steered minus baseline, percentage points) for Qwen2.5-7B-Instruct at Layer~25 with $\beta=1$ using the cultural prompt. Steering produces substantially different effects across language-region pairs, including both strong gains and degradations.}
\end{figure}


\newpage
% ...
\begin{figure}[t]
\subsection{Aya Expanse 8B}

    \centering
    \includegraphics[width=0.9\linewidth]{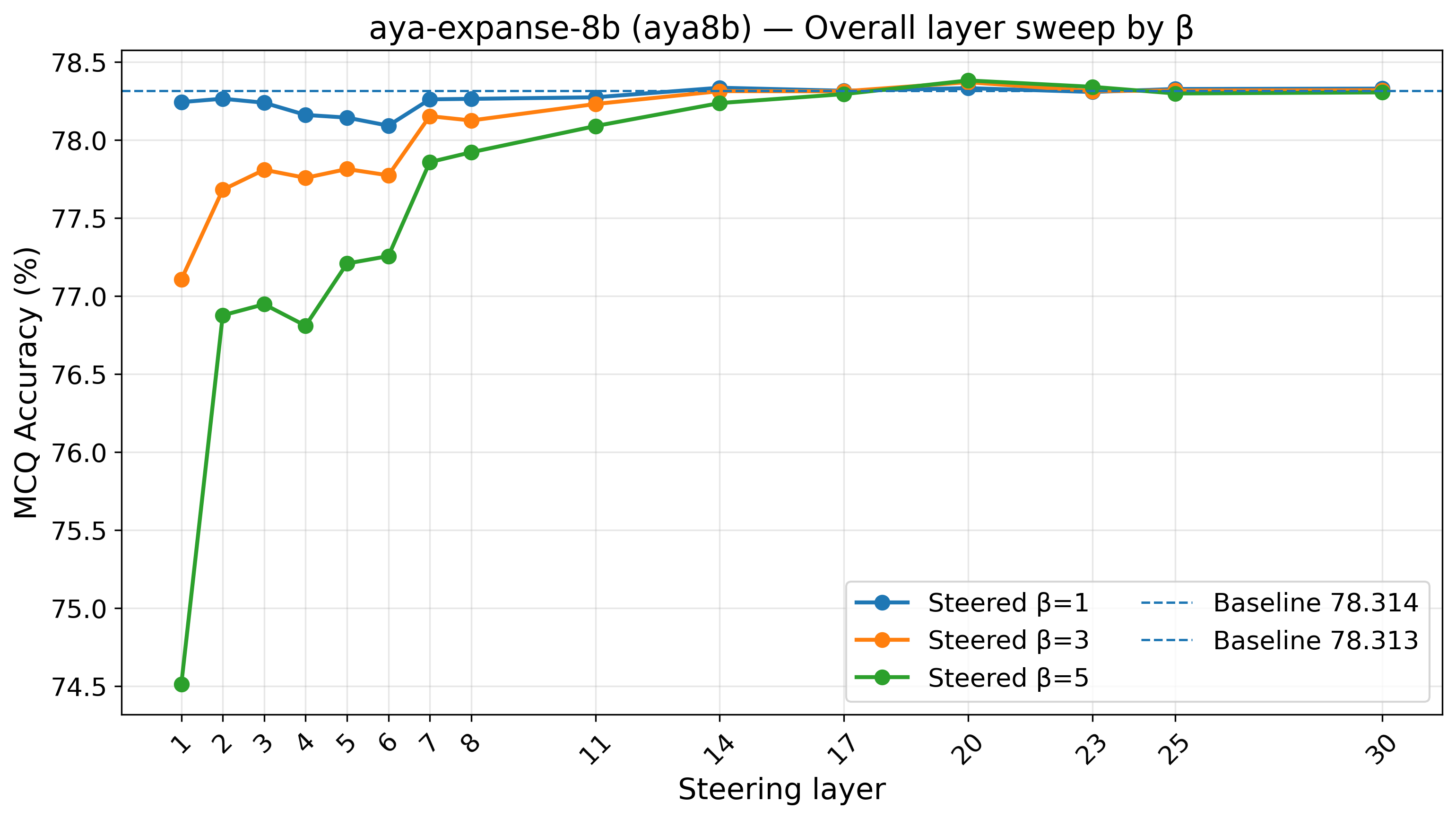}
    \caption{Post-hoc overall MCQ layer sweeps for Aya Expanse 8B under different steering strengths ($\beta \in \{1,3,5\}$). Large steering strengths can substantially degrade performance in early layers, while $\beta=1$ remains stable and yields the best overall trade-off in our experiments.}
\end{figure}


\begin{figure}[t]
    \centering
    \includegraphics[width=0.9\linewidth]{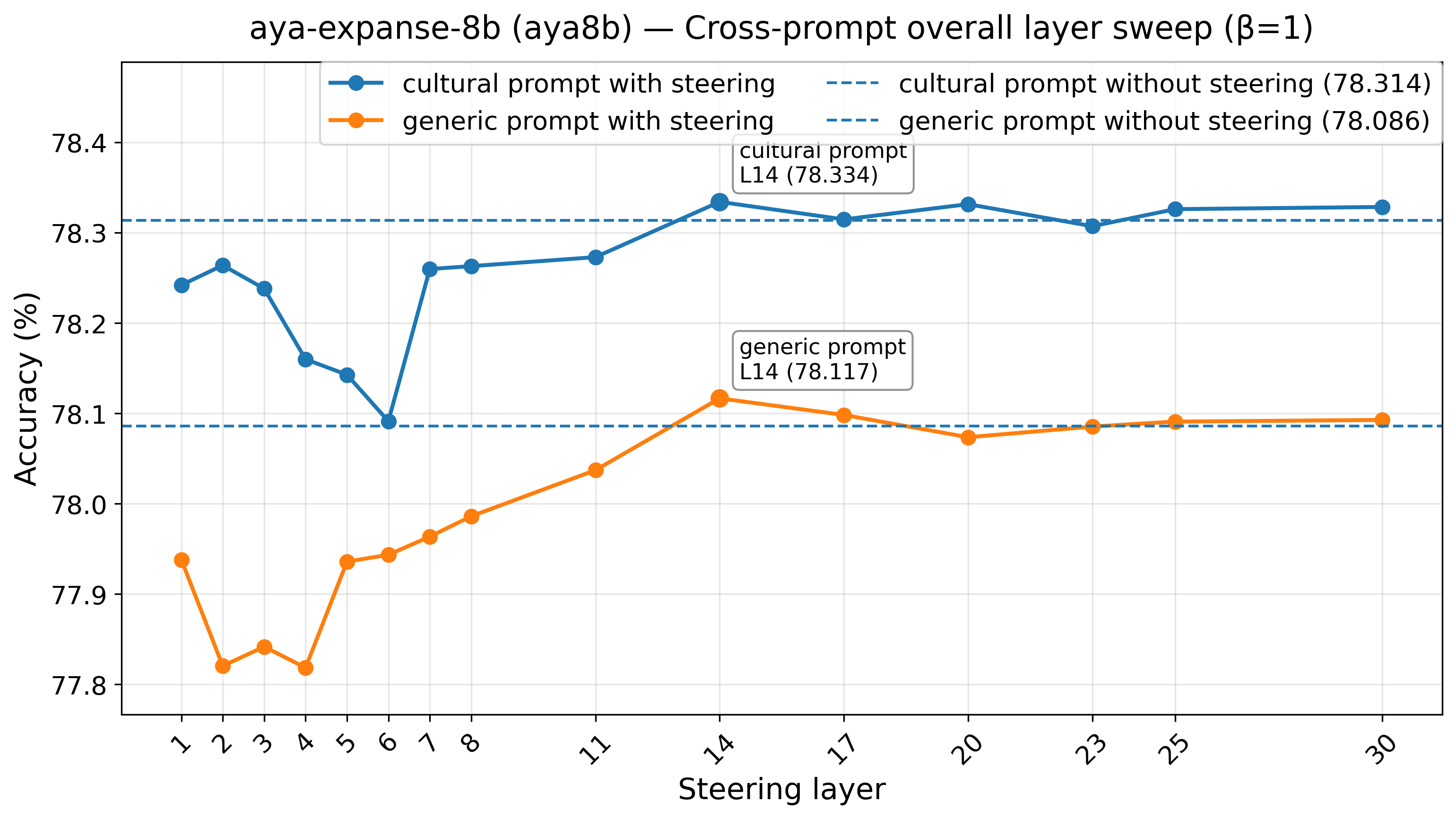}
    \caption{Post-hoc cross-prompt MCQ layer sweep for Aya Expanse 8B with $\beta=1$. The official submission uses the \textbf{cultural prompt}. Prompt choice affects the baseline accuracy but, with this model, delivers the same optimal steering layer (Layer~14 for both the cultural and generic prompt).}
\end{figure}


\begin{figure}[t]
    \centering
    \includegraphics[width=0.9\linewidth]{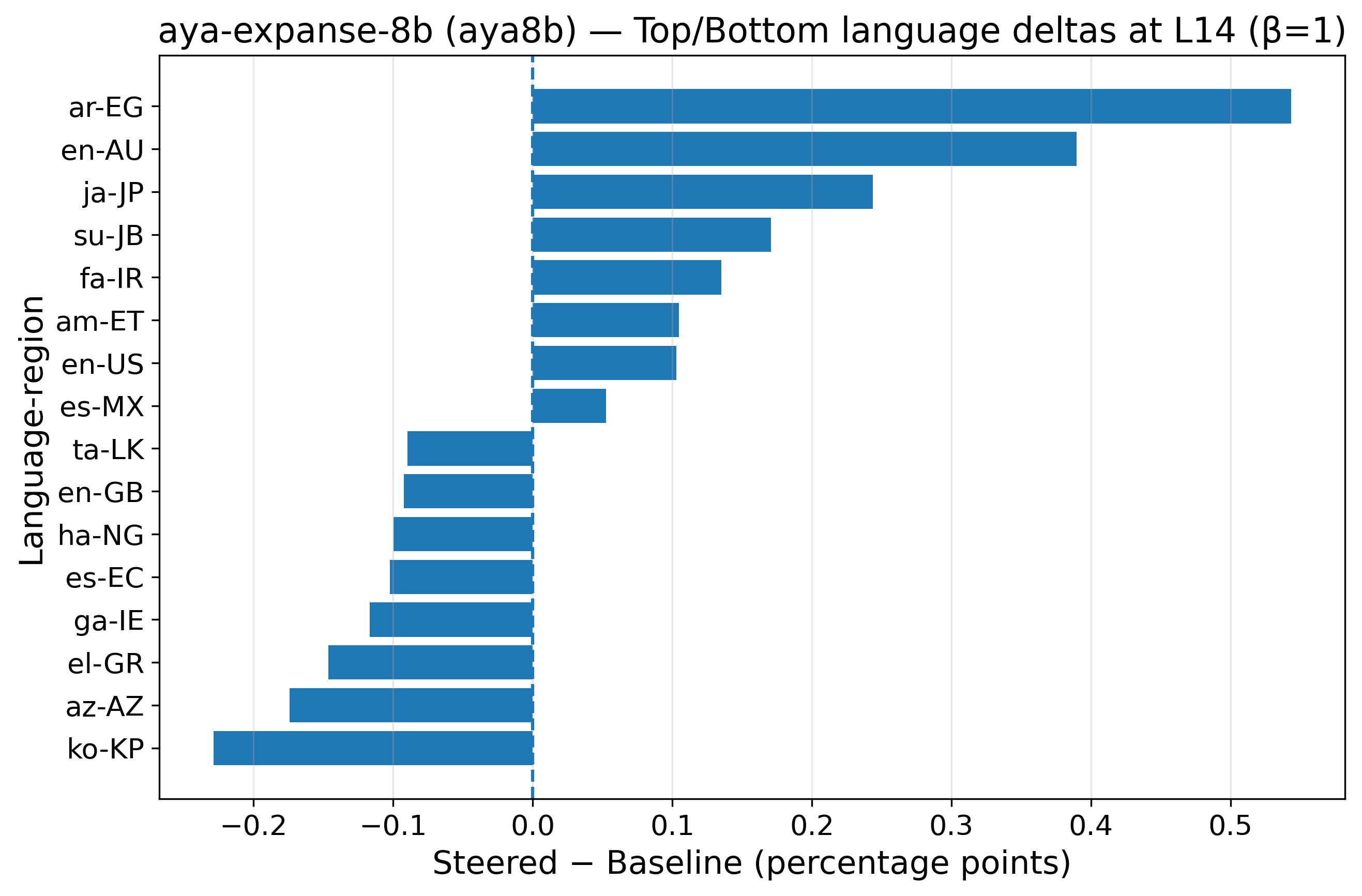}
    \caption{Top and bottom per-language MCQ accuracy changes (steered minus baseline, percentage points) for Aya Expanse 8B at Layer~14 with $\beta=1$ using the cultural prompt. Steering produces substantially different effects across language-region pairs, including both strong gains and degradations.}
\end{figure}


\newpage
% ...
\begin{figure}[t]
\subsection{Aya Expanse 32B}

    \centering
    \includegraphics[width=0.9\linewidth]{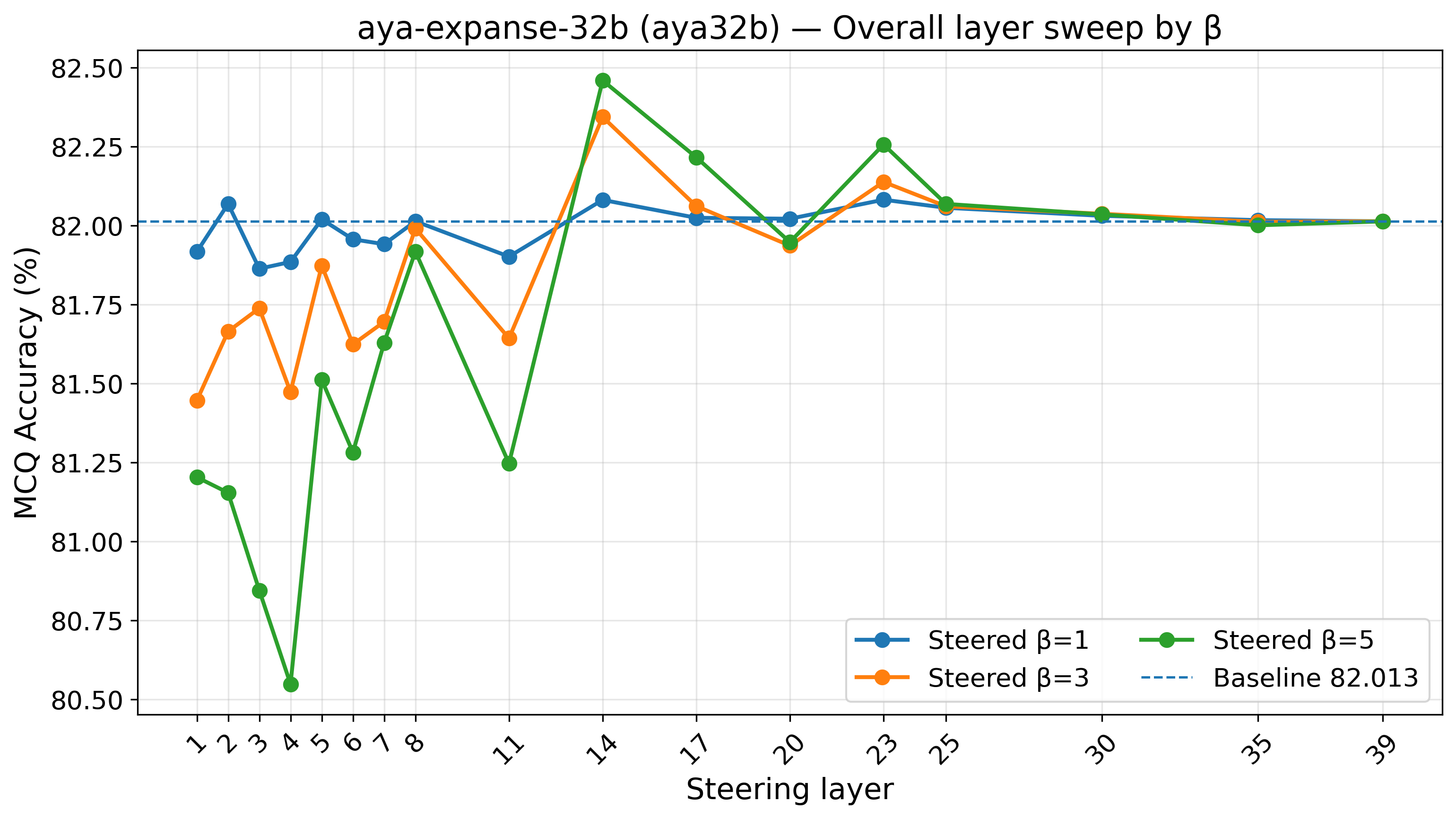}
    \caption{Post-hoc overall MCQ layer sweeps for Aya Expanse 32B under different steering strengths ($\beta \in \{1,3,5\}$). Large steering strengths can substantially degrade performance in early layers, while improving or meeting performance in mid to late layers. In comparison, $\beta=1$ remains stable across layers although $\beta=5$ yields the best overall Acc in our experiments.}
\end{figure}


\begin{figure}[t]
    \centering
    \includegraphics[width=0.9\linewidth]{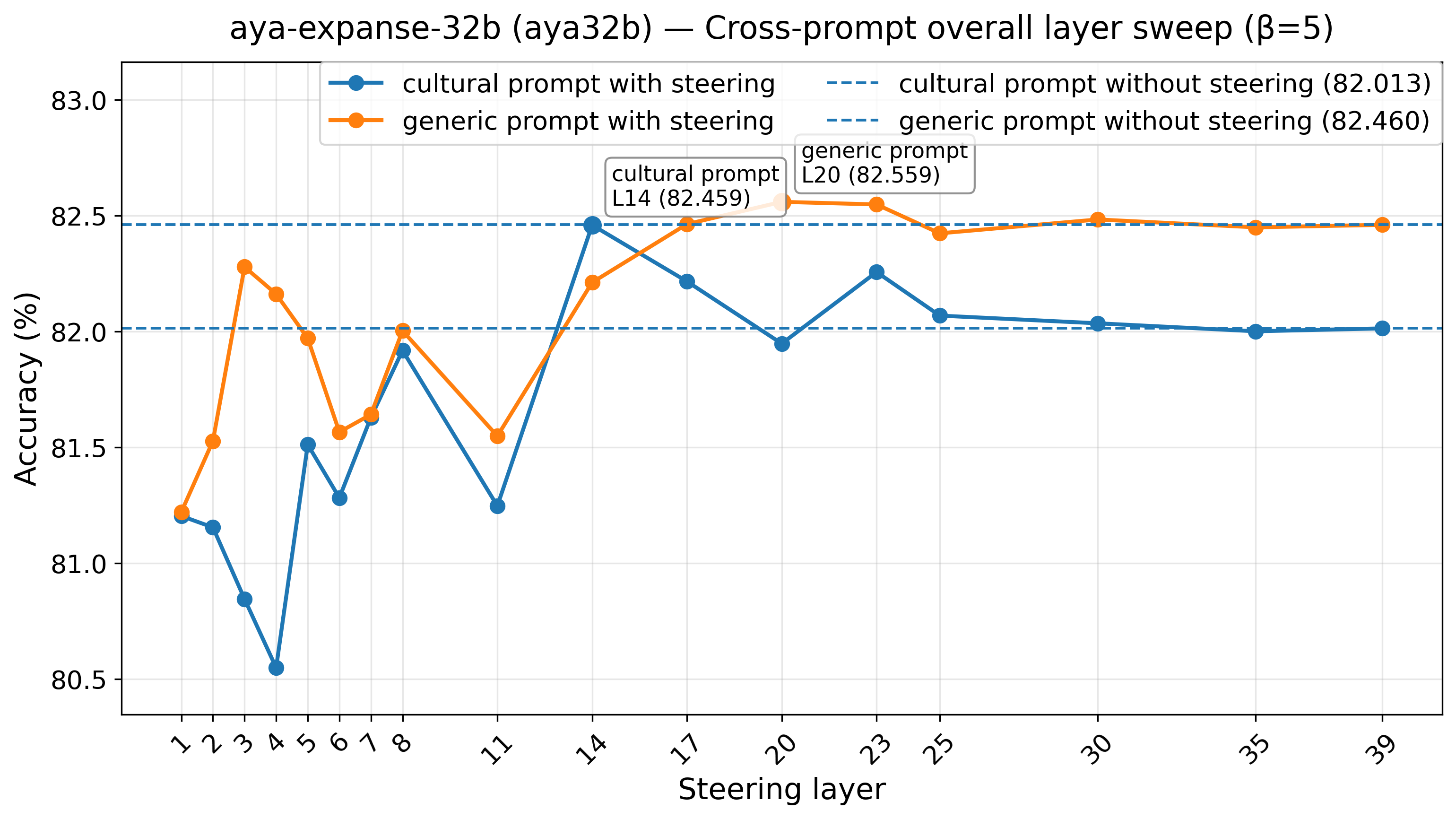}
    \caption{Post-hoc cross-prompt MCQ layer sweep for Aya Expanse 32B with $\beta=5$. The official submission uses the \textbf{cultural prompt}. Prompt choice affects both baseline accuracy and the optimal steering layer (here, Layer~14 for the cultural prompt and Layer~20 for the generic prompt).}
\end{figure}


\begin{figure}[t]
    \centering
    \includegraphics[width=0.9\linewidth]{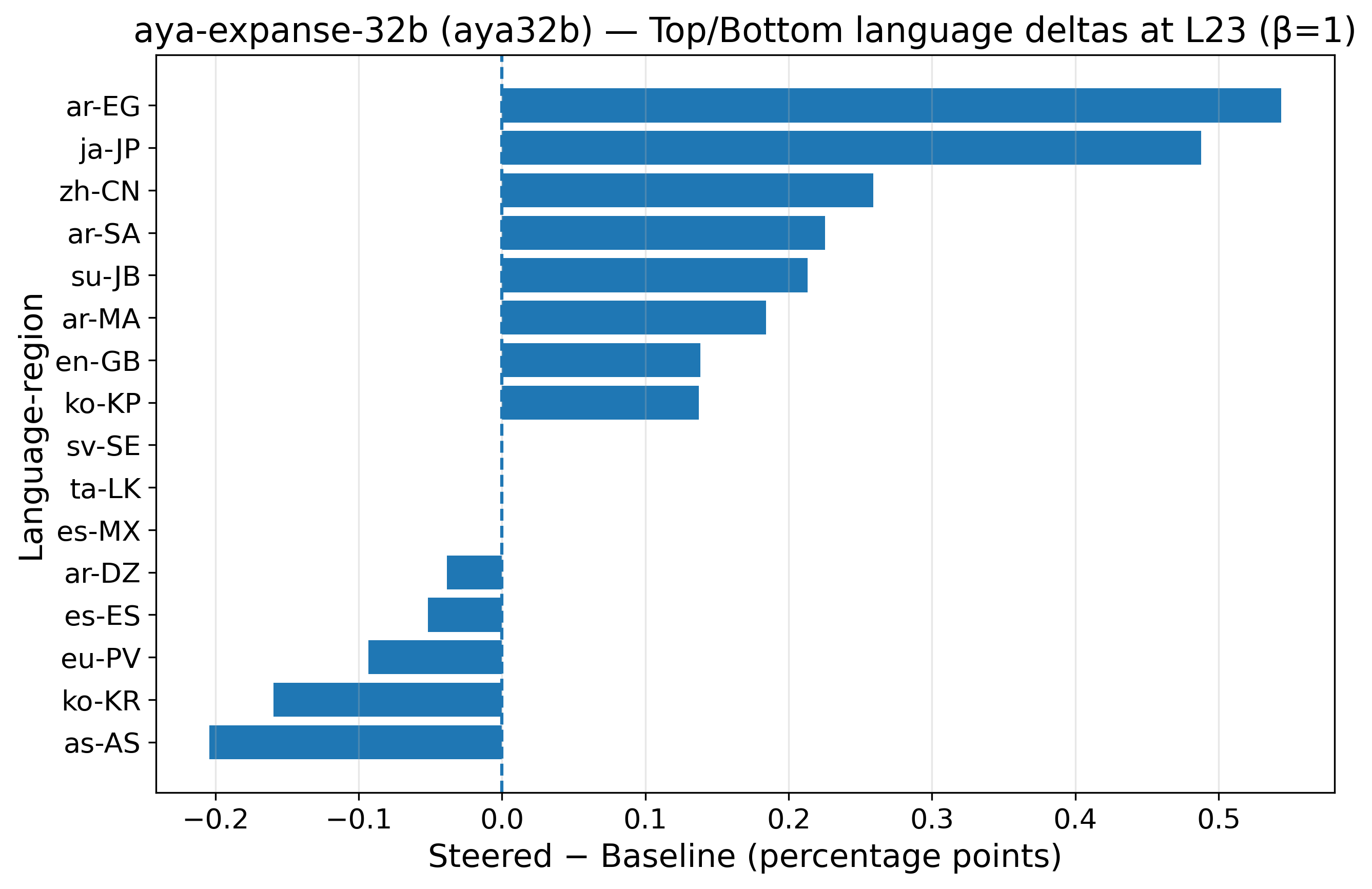}
    \caption{Top and bottom per-language MCQ accuracy changes (steered minus baseline, percentage points) for Aya Expanse 32B at Layer~23 with $\beta=1$ using the cultural prompt. Steering produces substantially different effects across language-region pairs, including both strong gains and degradations.}
\end{figure}


\newpage
% ...
\begin{figure}[t]
\subsection{Qwen3-8B}
    \centering
    \includegraphics[width=0.9\linewidth]{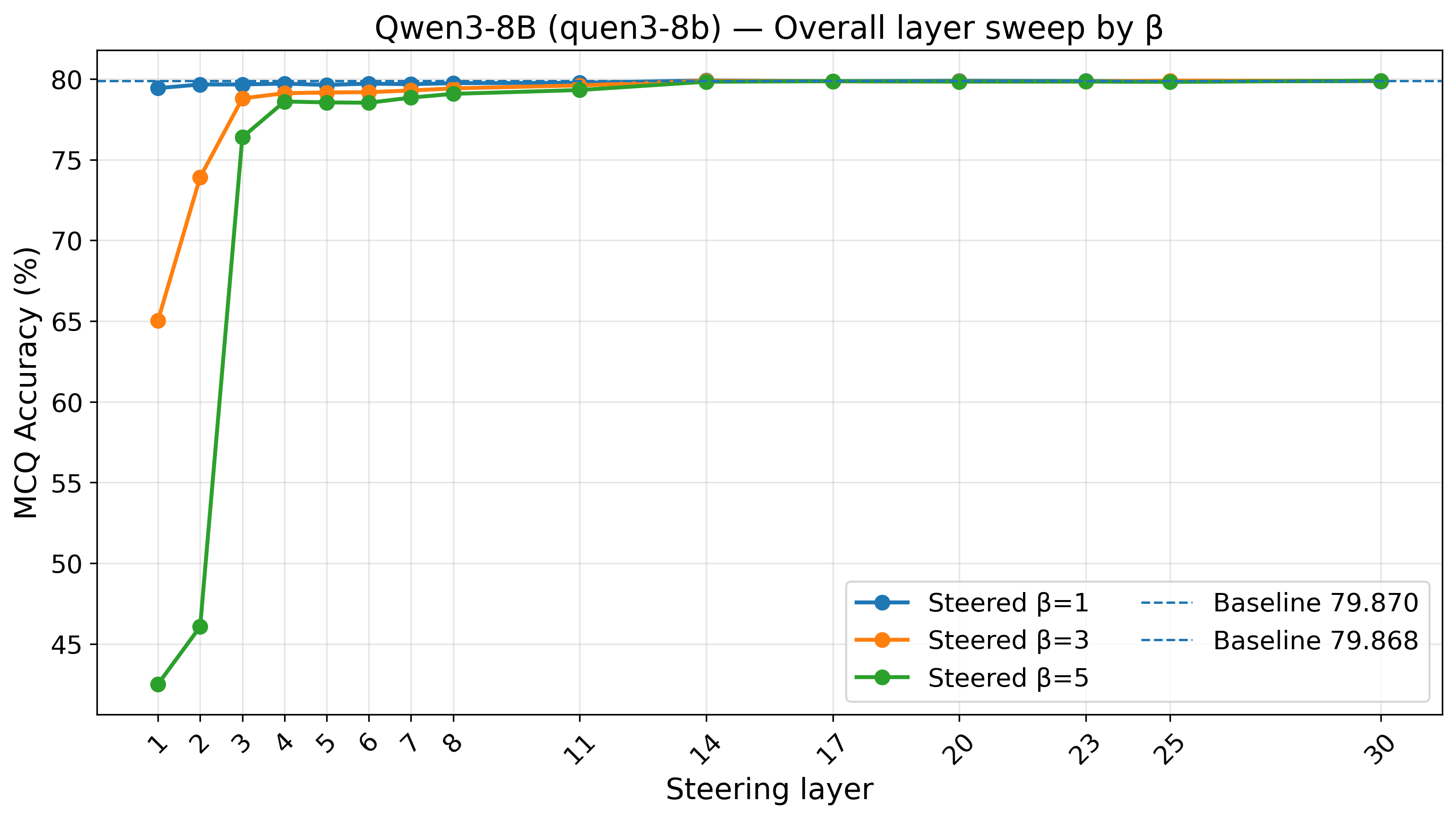}
    \caption{Post-hoc overall MCQ layer sweeps for Qwen3-8B under different steering strengths ($\beta \in \{1,3,5\}$). Large steering strengths can substantially degrade performance in early layers, while $\beta=1$ remains stable and yields the best overall trade-off in our experiments.}
\end{figure}


\begin{figure}[t]
    \centering
    \includegraphics[width=0.9\linewidth]{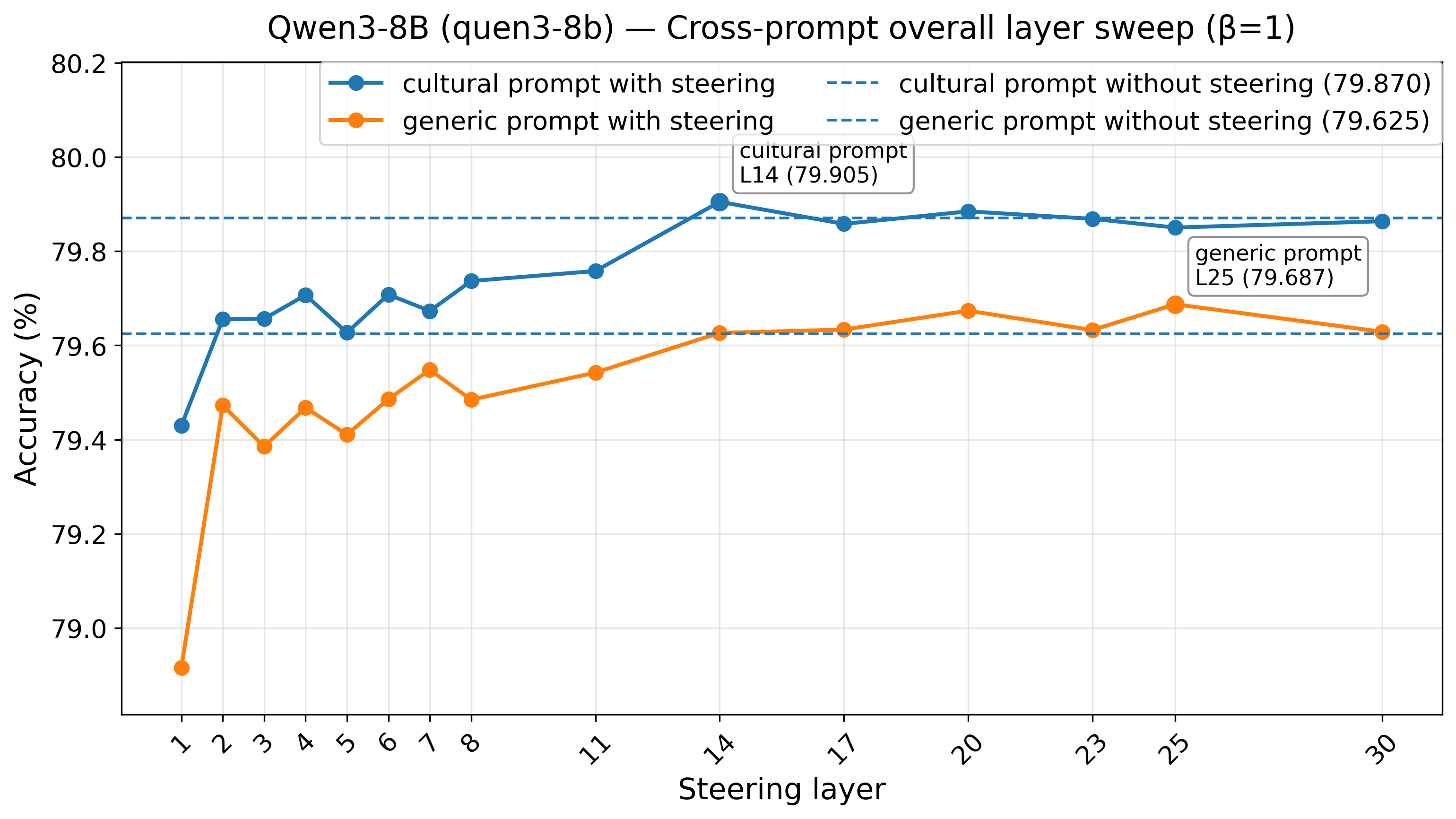}
    \caption{Post-hoc cross-prompt MCQ layer sweep for Qwen3-8B with $\beta=1$. The official submission uses the \textbf{cultural prompt}. Prompt choice affects both baseline accuracy and the optimal steering layer (here, Layer~14 for the cultural prompt and Layer~25 for the generic prompt).}
\end{figure}


\begin{figure}[t]
    \centering
    \includegraphics[width=0.9\linewidth]{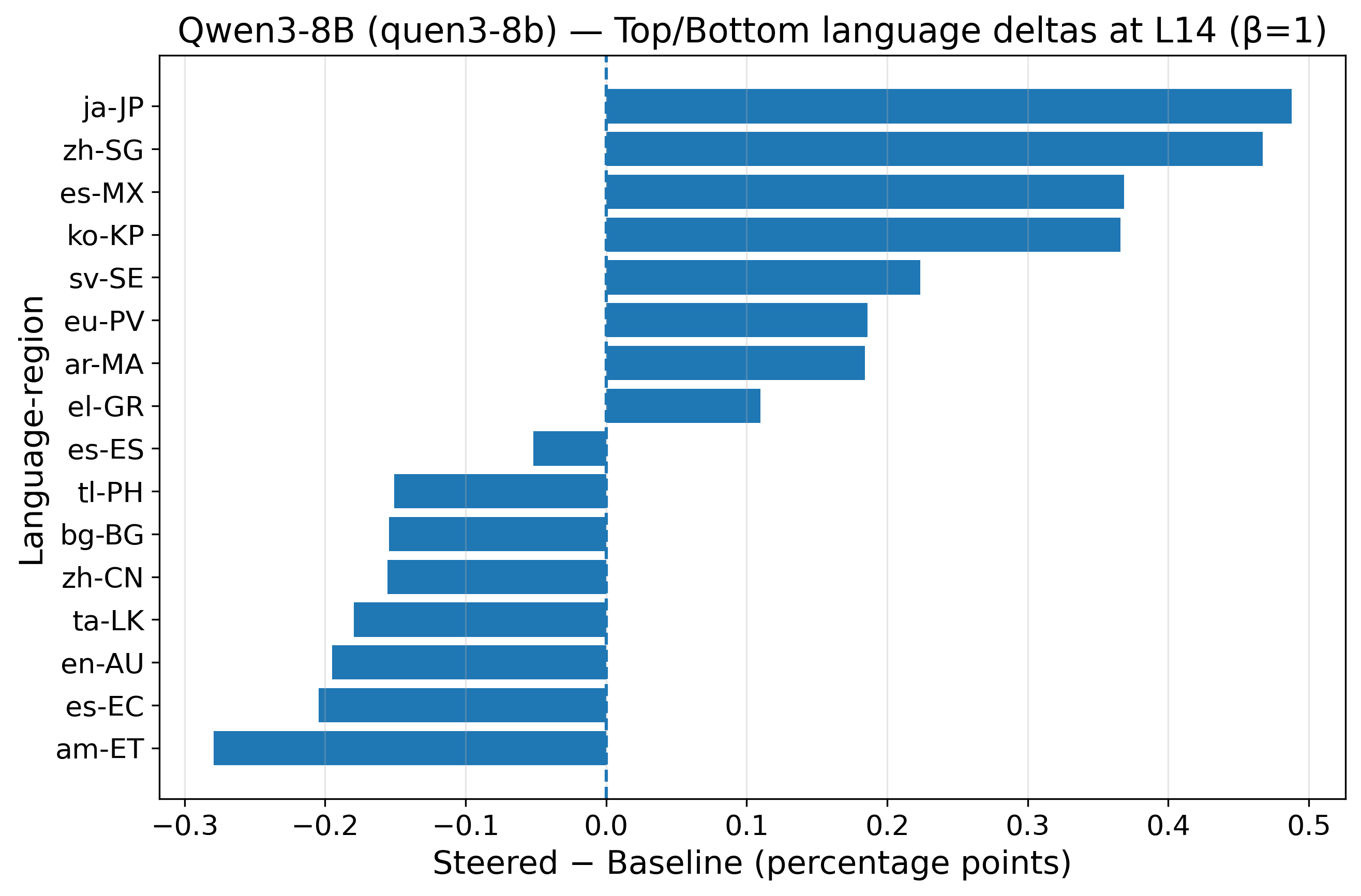}
    \caption{Top and bottom per-language MCQ accuracy changes (steered minus baseline, percentage points) for Qwen3-8B at Layer~14 with $\beta=1$ using the cultural prompt. Steering produces substantially different effects across language-region pairs, including both strong gains and degradations.}
\end{figure}


% ...
\newpage
% ...
\begin{figure}[t]
\subsection{Qwen3-32B}
    \centering
    \includegraphics[width=0.9\linewidth]{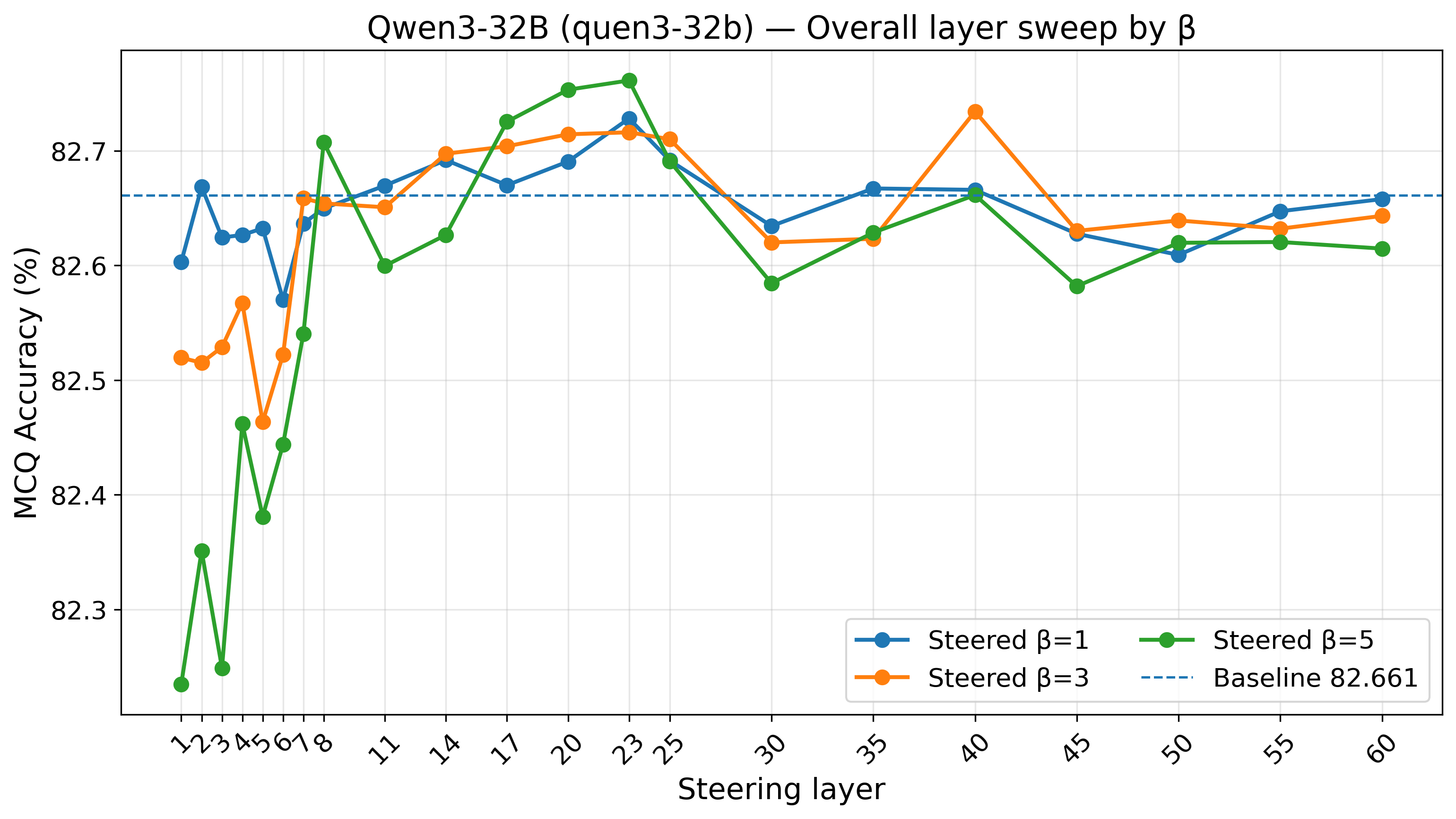}
    \caption{Post-hoc overall MCQ layer sweeps for Qwen3-32B under different steering strengths ($\beta \in \{1,3,5\}$). Large steering strengths can substantially degrade performance in early layers, while $\beta=1$ remains stable although $\beta=5$ yields the best overall Acc in our experiments.}
\end{figure}


\begin{figure}[t]
    \centering
    \includegraphics[width=0.9\linewidth]{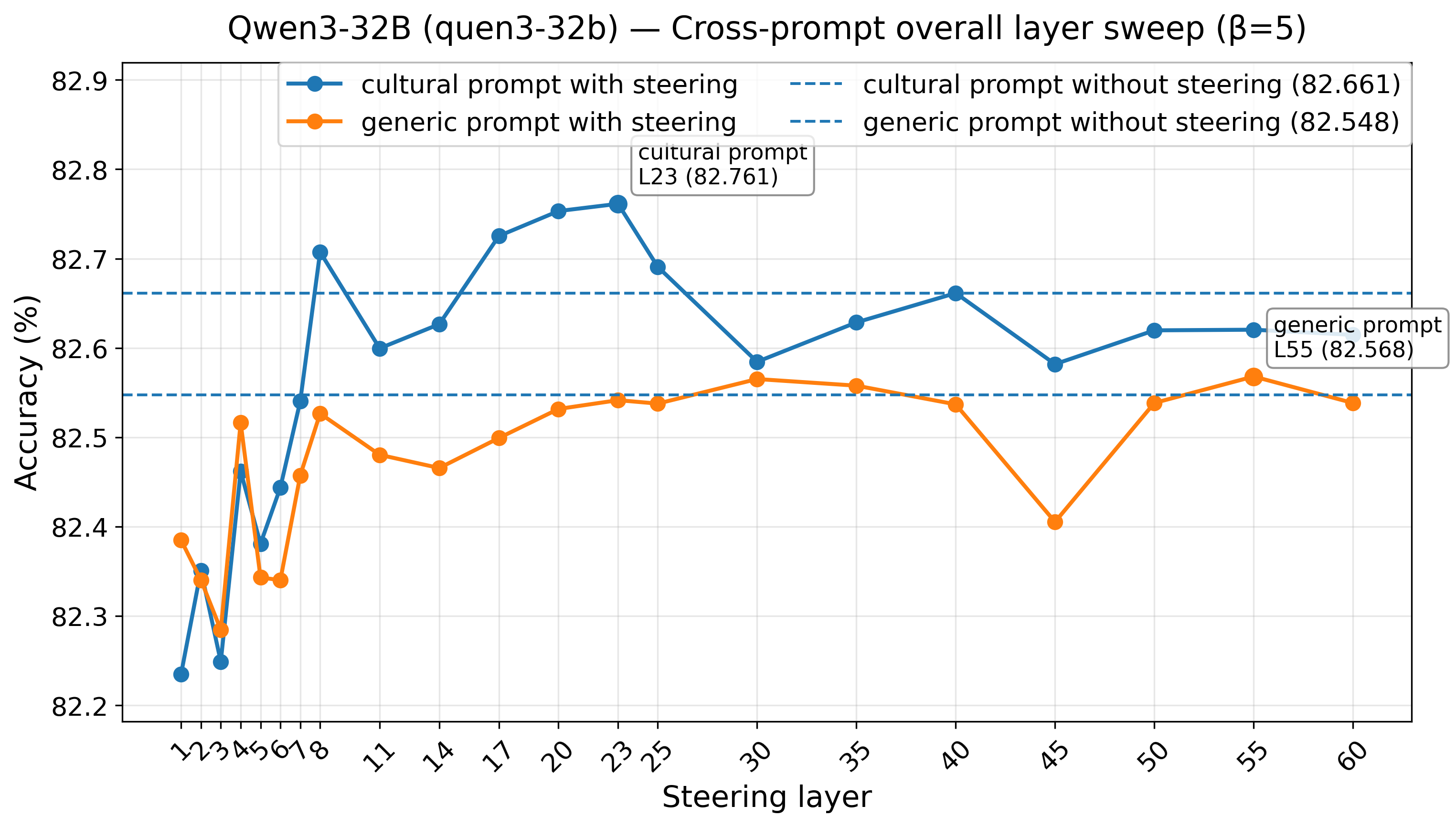}
    \caption{Post-hoc cross-prompt MCQ layer sweep for Qwen3-32B with $\beta=5$. The official submission uses the \textbf{cultural prompt}. Prompt choice affects both baseline accuracy and the optimal steering layer (here, Layer~23 for the cultural prompt and Layer~55 for the generic prompt).}
\end{figure}


\begin{figure}[t]
    \centering
    \includegraphics[width=0.9\linewidth]{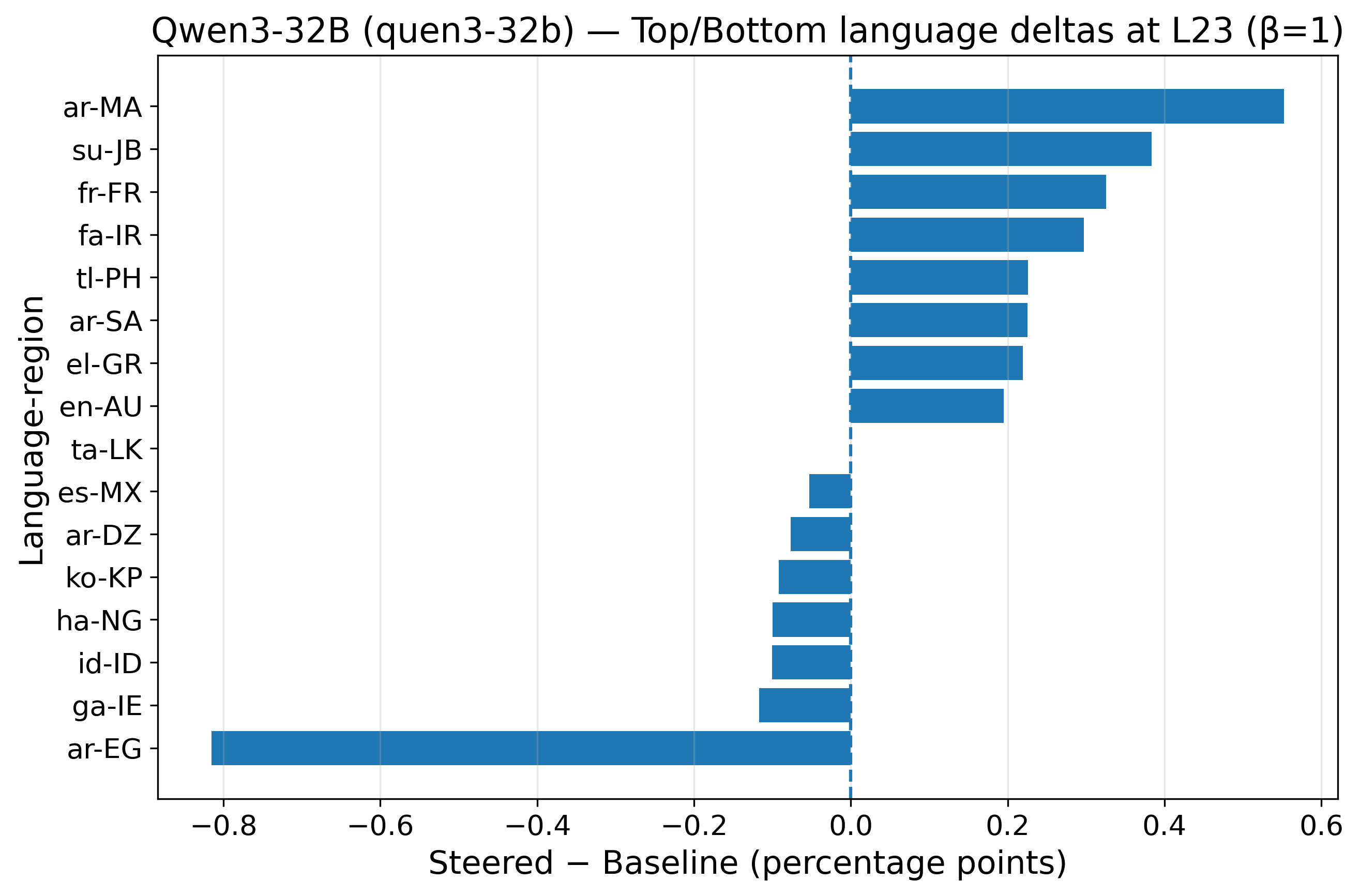}
    \caption{Top and bottom per-language MCQ accuracy changes (steered minus baseline, percentage points) for Qwen3-32B at Layer~23 with $\beta=1$ using the cultural prompt. Steering produces substantially different effects across language-region pairs, including both strong gains and degradations.}
\end{figure}


\newpage
\clearpage

\begin{figure}[t]
\section{Post-hoc SAQ Analysis Plots}
\label{app:posthoc-plots-saq}
This appendix provides additional post-hoc analysis plots for all tested models.
\subsection{Qwen2.5-72B-Instruct}
    \centering
    \includegraphics[width=0.9\linewidth]{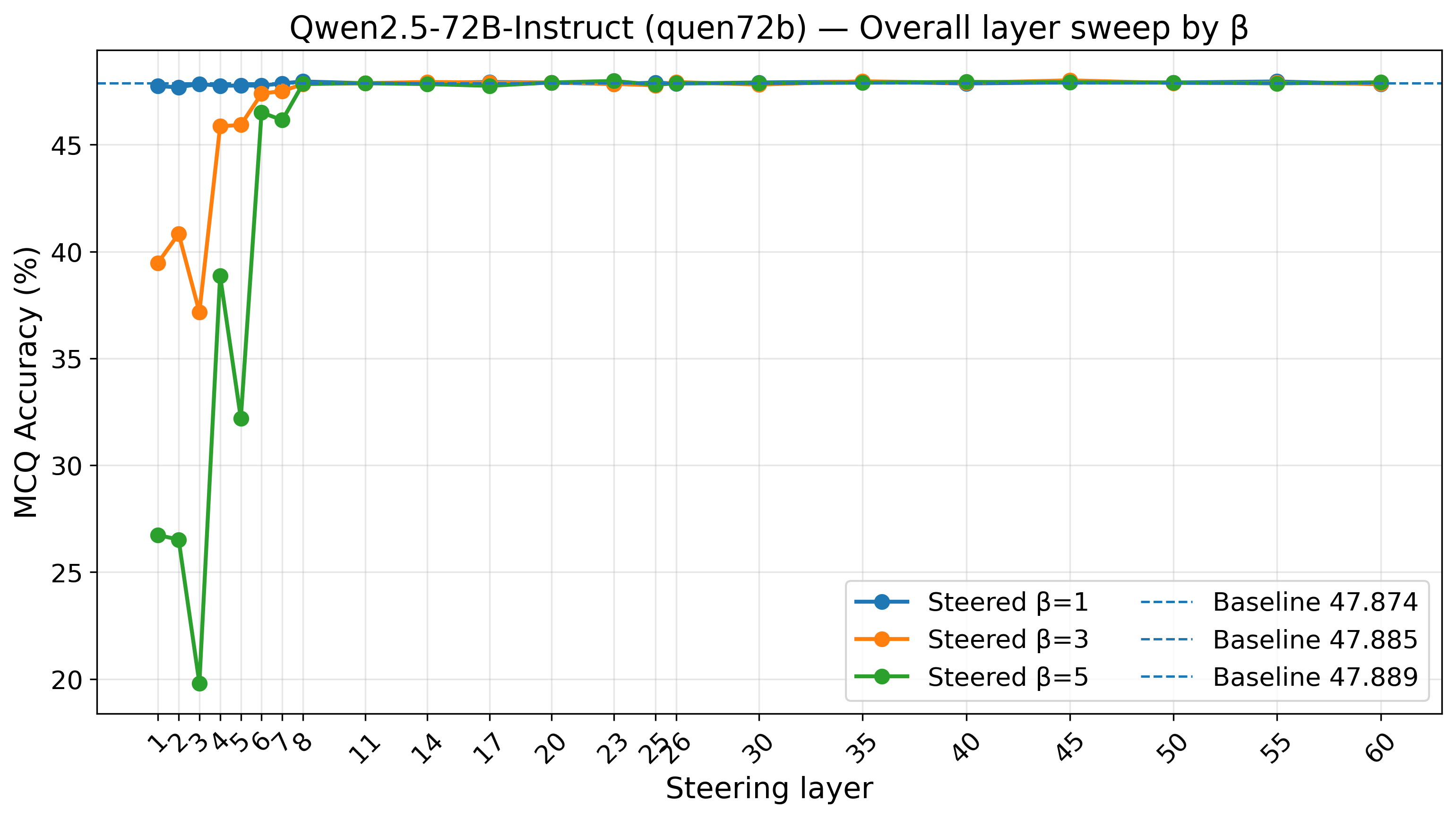}
    \caption{Post-hoc overall SAQ layer sweeps for Qwen2.5-72B-Instruct under different steering strengths ($\beta \in \{1,3,5\}$). Large steering strengths can substantially degrade performance in early layers, while $\beta=1$ remains stable and yields the best overall trade-off in our experiments.}
\end{figure}


\begin{figure}[t]
    \centering
    \includegraphics[width=0.9\linewidth]{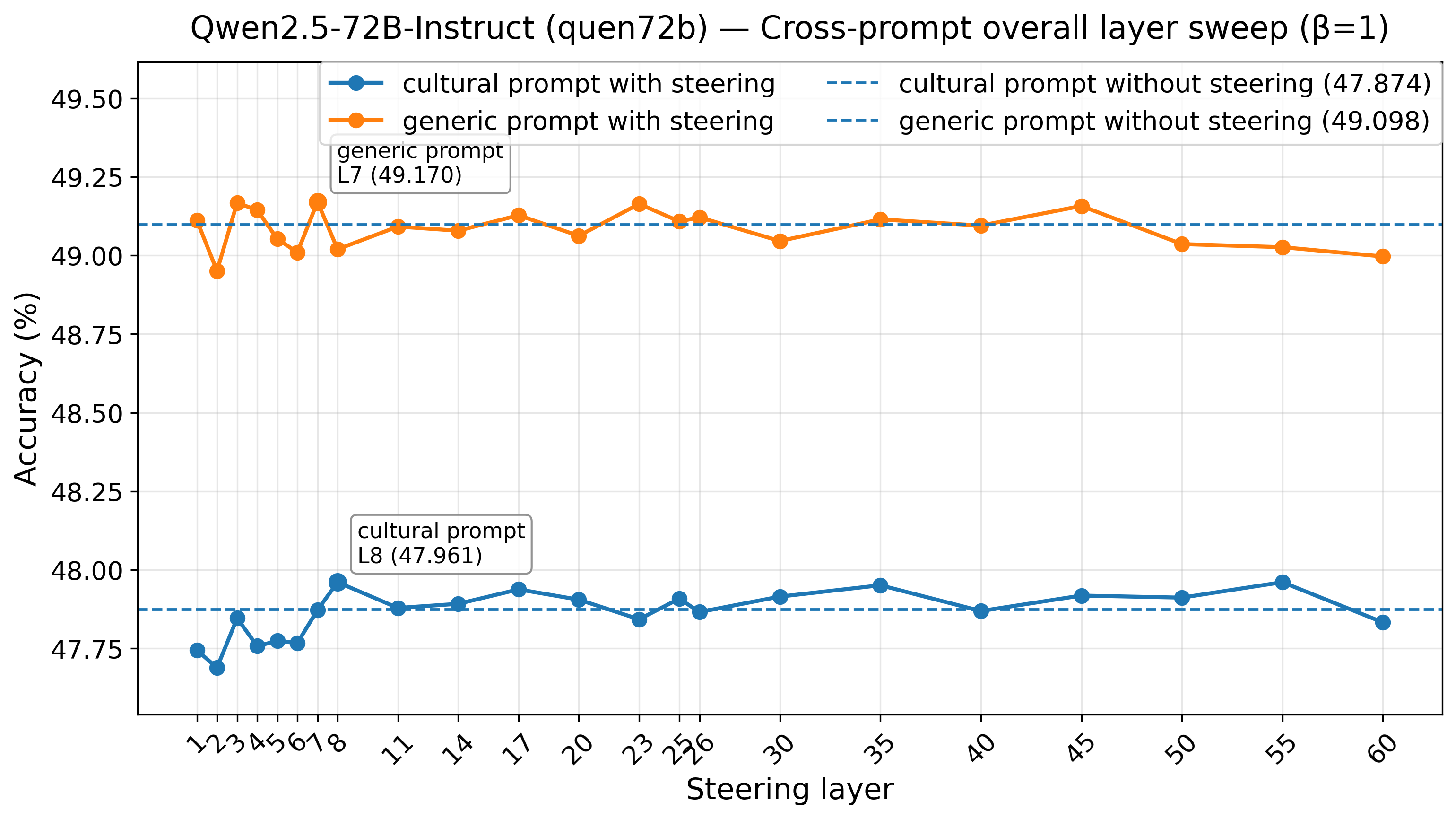}
    \caption{Post-hoc cross-prompt SAQ layer sweep for Qwen2.5-72B-Instruct with $\beta=1$. The official submission uses the \textbf{cultural prompt}. Prompt choice affects both baseline accuracy and the optimal steering layer (here, Layer~8 for the cultural prompt and Layer~7 for the generic prompt).}
\end{figure}


\begin{figure}[t]
    \centering
    \includegraphics[width=0.9\linewidth]{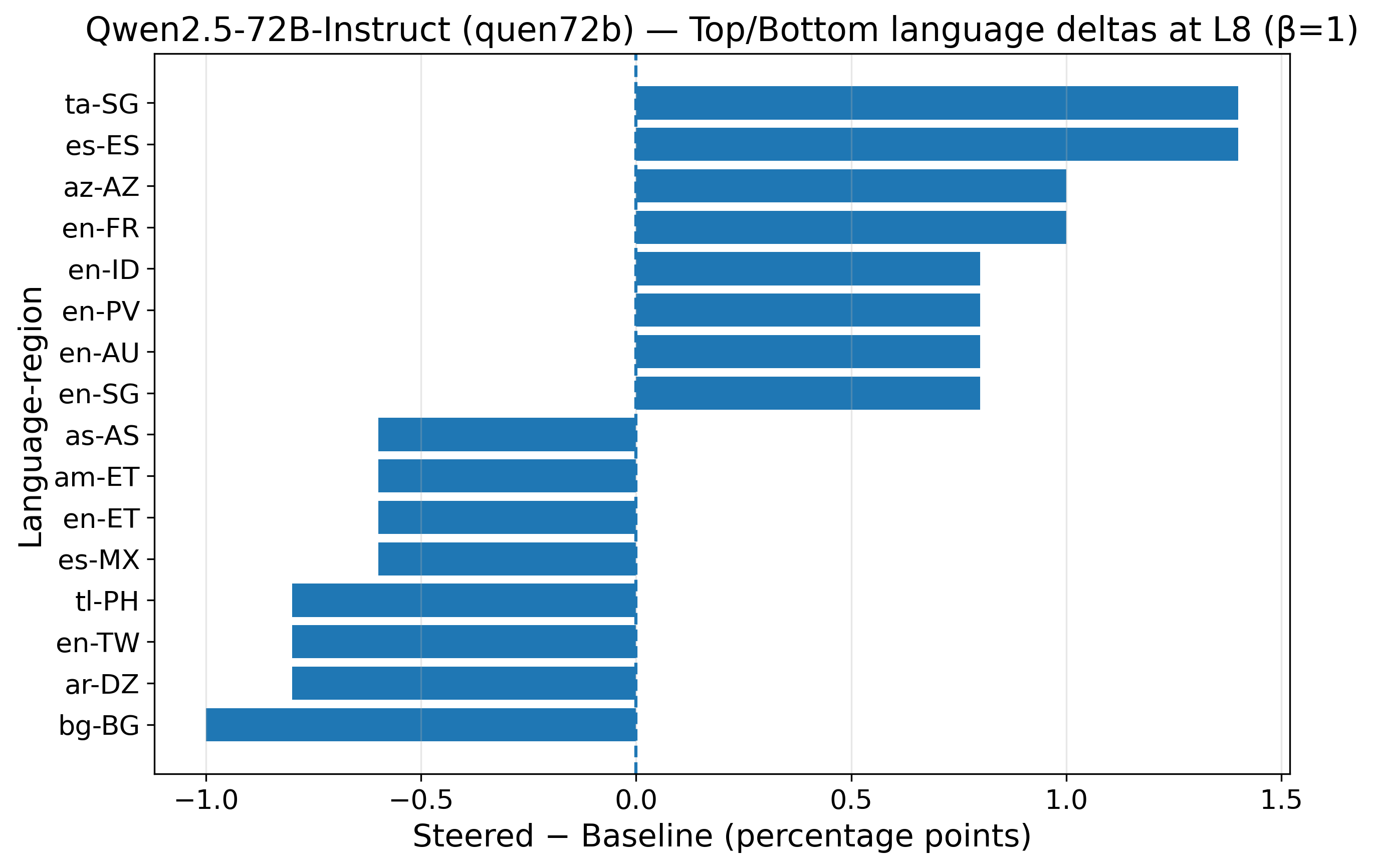}
    \caption{Top and bottom per-language SAQ accuracy changes (steered minus baseline, percentage points) for Qwen2.5-72B-Instruct at Layer~8 with $\beta=1$ using the cultural prompt. Steering produces substantially different effects across language-region pairs, including both strong gains and degradations.}
\end{figure}



\newpage
\begin{figure}[t]
\subsection{Qwen2.5-7B-Instruct}

    \centering
    \includegraphics[width=0.9\linewidth]{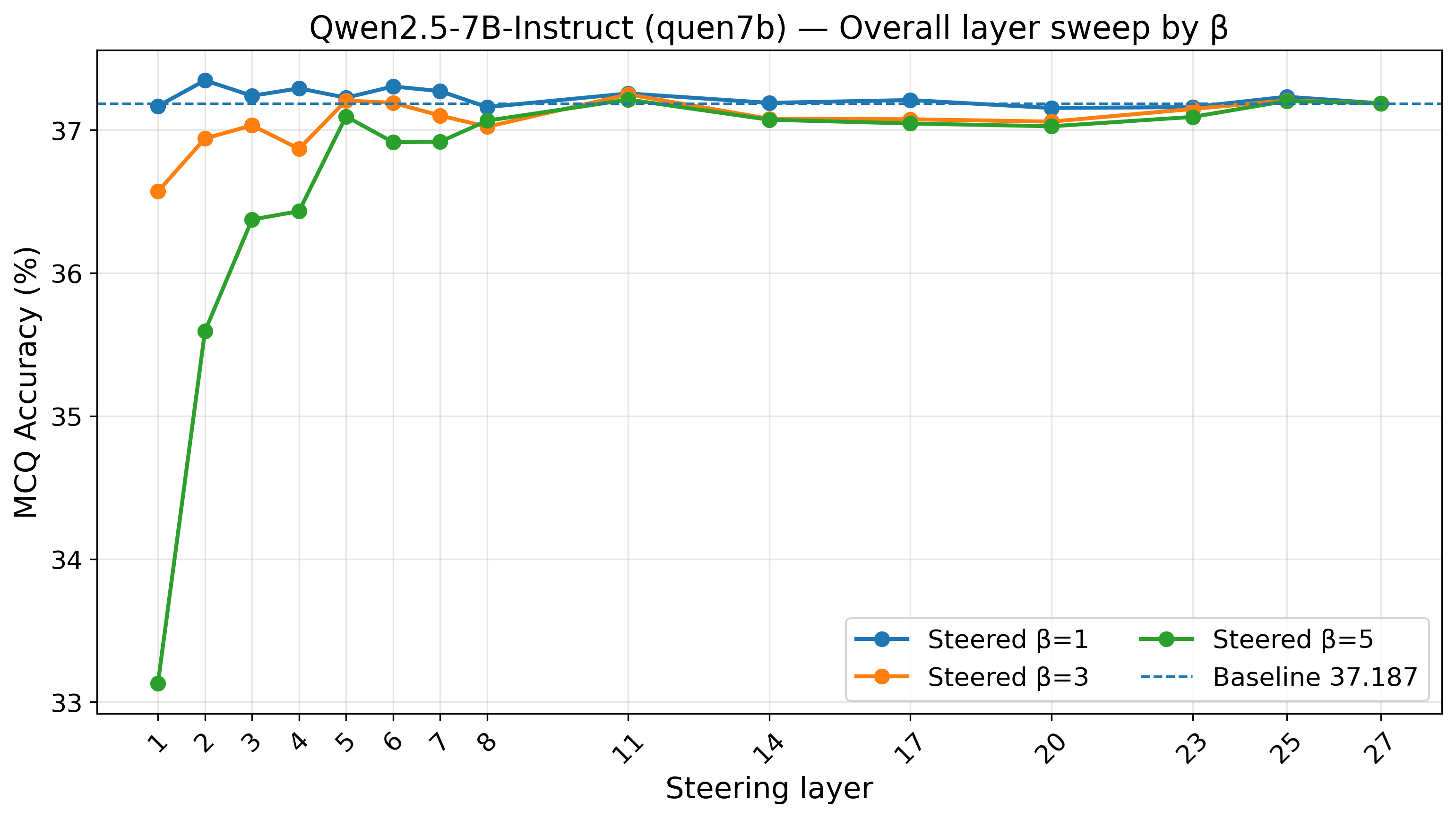}
    \caption{Post-hoc overall SAQ layer sweeps for Qwen2.5-7B-Instruct under different steering strengths ($\beta \in \{1,3,5\}$). Large steering strengths can substantially degrade performance in early layers, while $\beta=1$ remains stable and yields the best overall trade-off in our experiments.}
\end{figure}


\begin{figure}[t]
    \centering
    \includegraphics[width=0.9\linewidth]{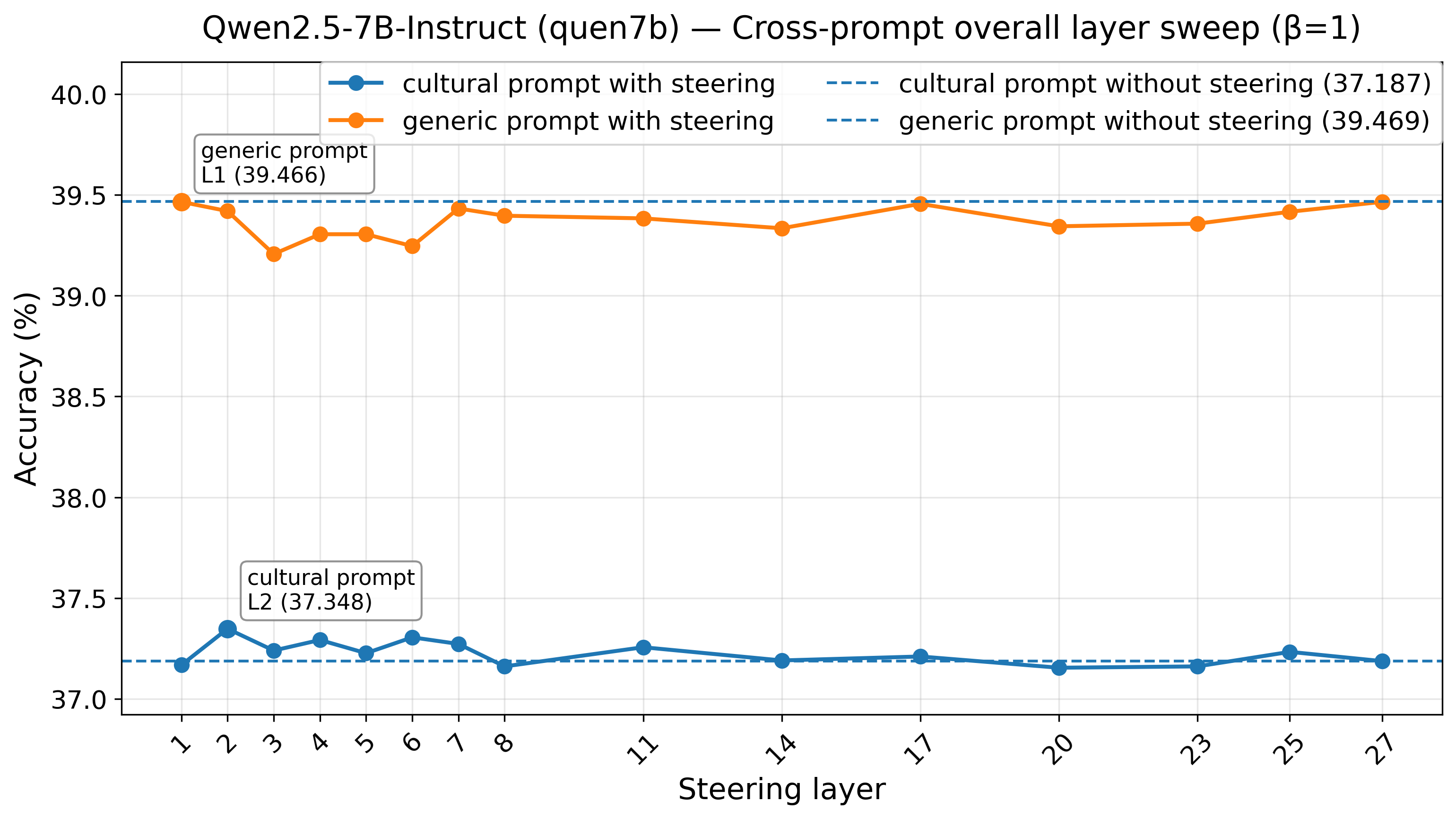}
    \caption{Post-hoc cross-prompt SAQ layer sweep for Qwen2.5-7B-Instruct with $\beta=1$. The official submission uses the \textbf{cultural prompt}. Prompt choice affects both baseline accuracy and the optimal steering layer (here, Layer~2 for the cultural prompt and Layer~1 for the generic prompt).}
\end{figure}


\begin{figure}[t]
    \centering
    \includegraphics[width=0.9\linewidth]{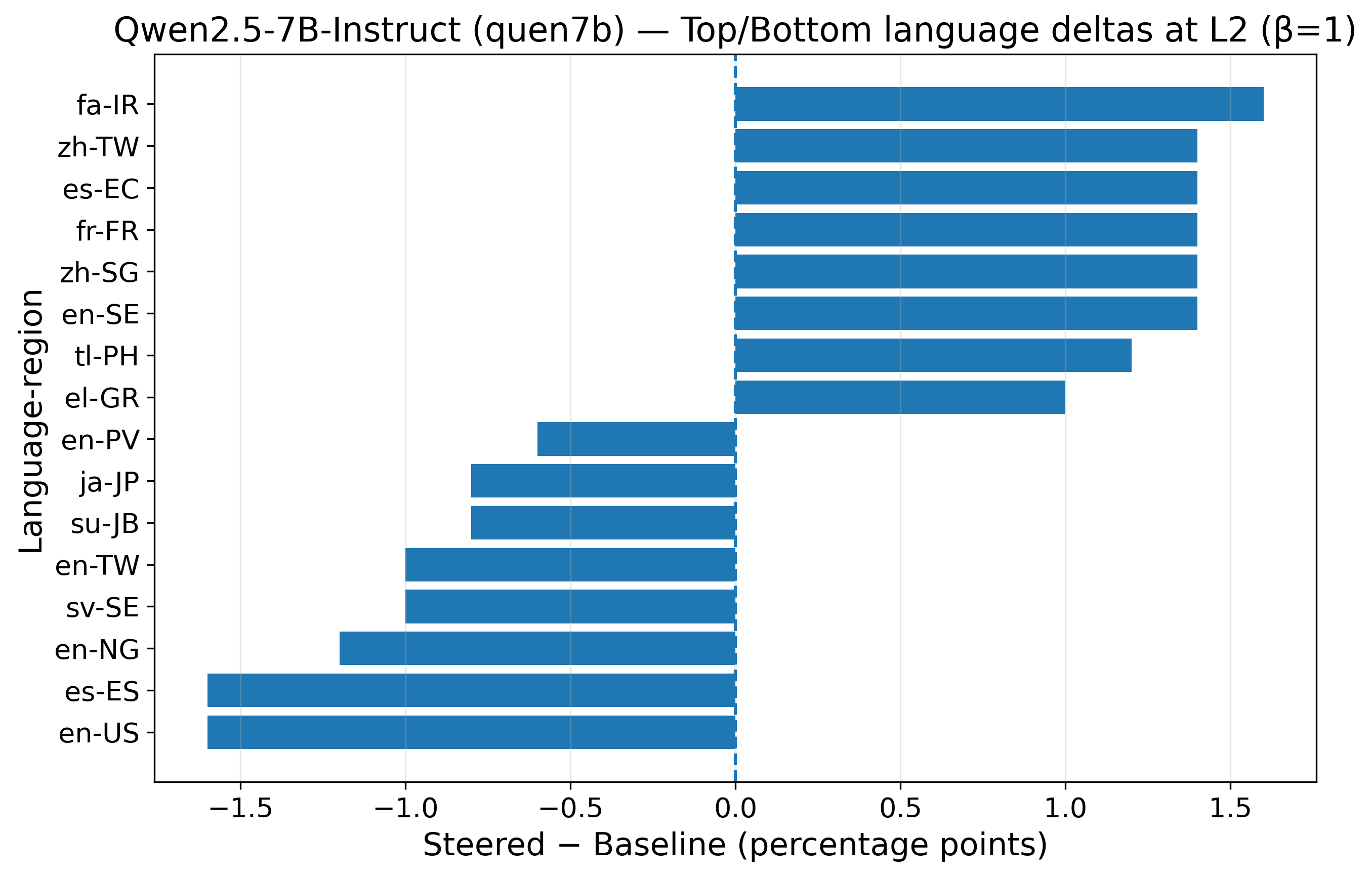}
    \caption{Top and bottom per-language SAQ accuracy changes (steered minus baseline, percentage points) for Qwen2.5-7B-Instruct at Layer~2 with $\beta=1$ using the cultural prompt. Steering produces substantially different effects across language-region pairs, including both strong gains and degradations.}
\end{figure}


\newpage
% ...
\begin{figure}[t]
\subsection{Aya Expanse 8B}

    \centering
    \includegraphics[width=0.9\linewidth]{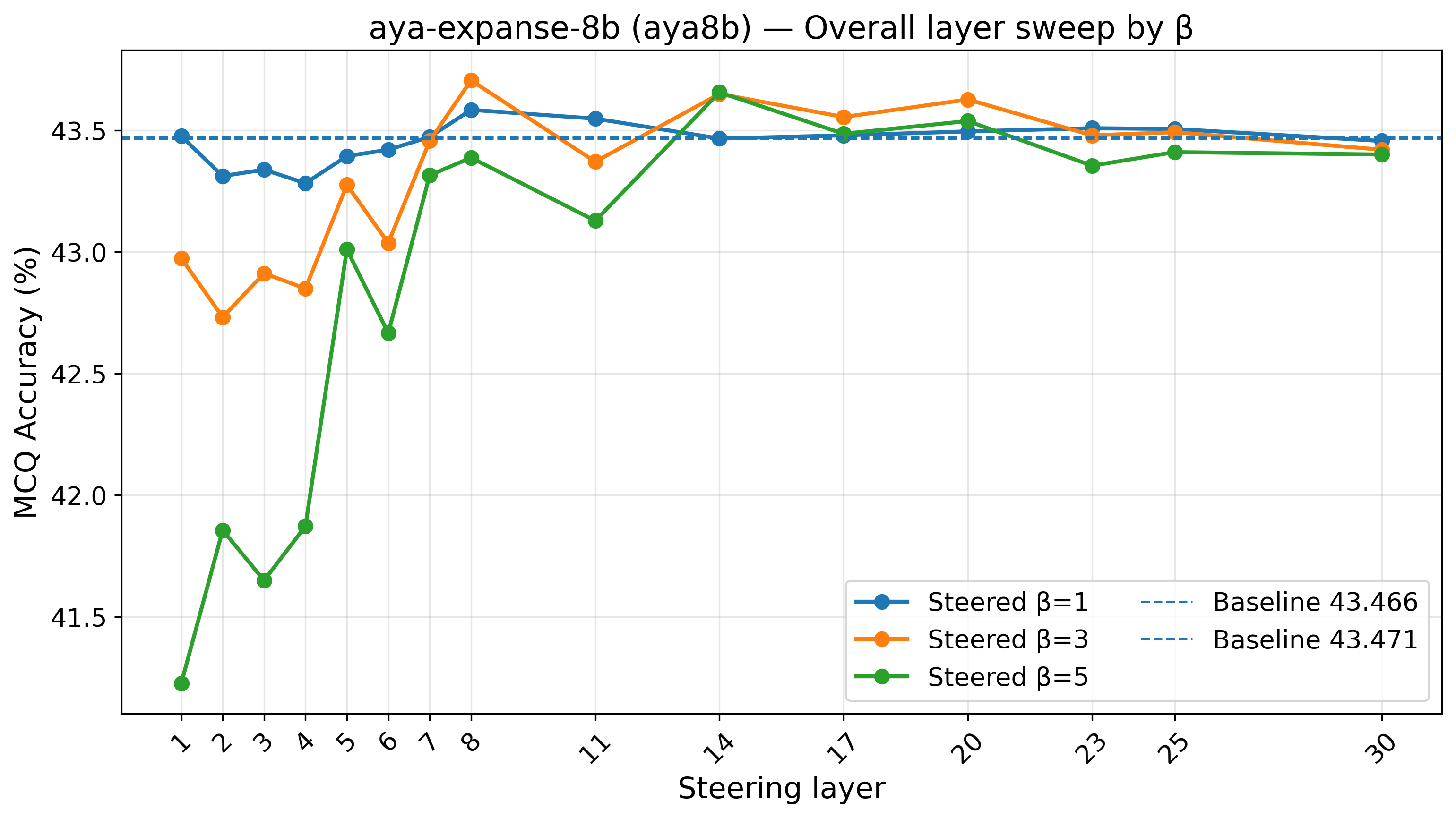}
    \caption{Post-hoc overall SAQ layer sweeps for Aya Expanse 8B under different steering strengths ($\beta \in \{1,3,5\}$). Large steering strengths can substantially degrade performance in early layers, while $\beta=1$ remains stable and yields the best overall trade-off in our experiments.}
\end{figure}


\begin{figure}[t]
    \centering
    \includegraphics[width=0.9\linewidth]{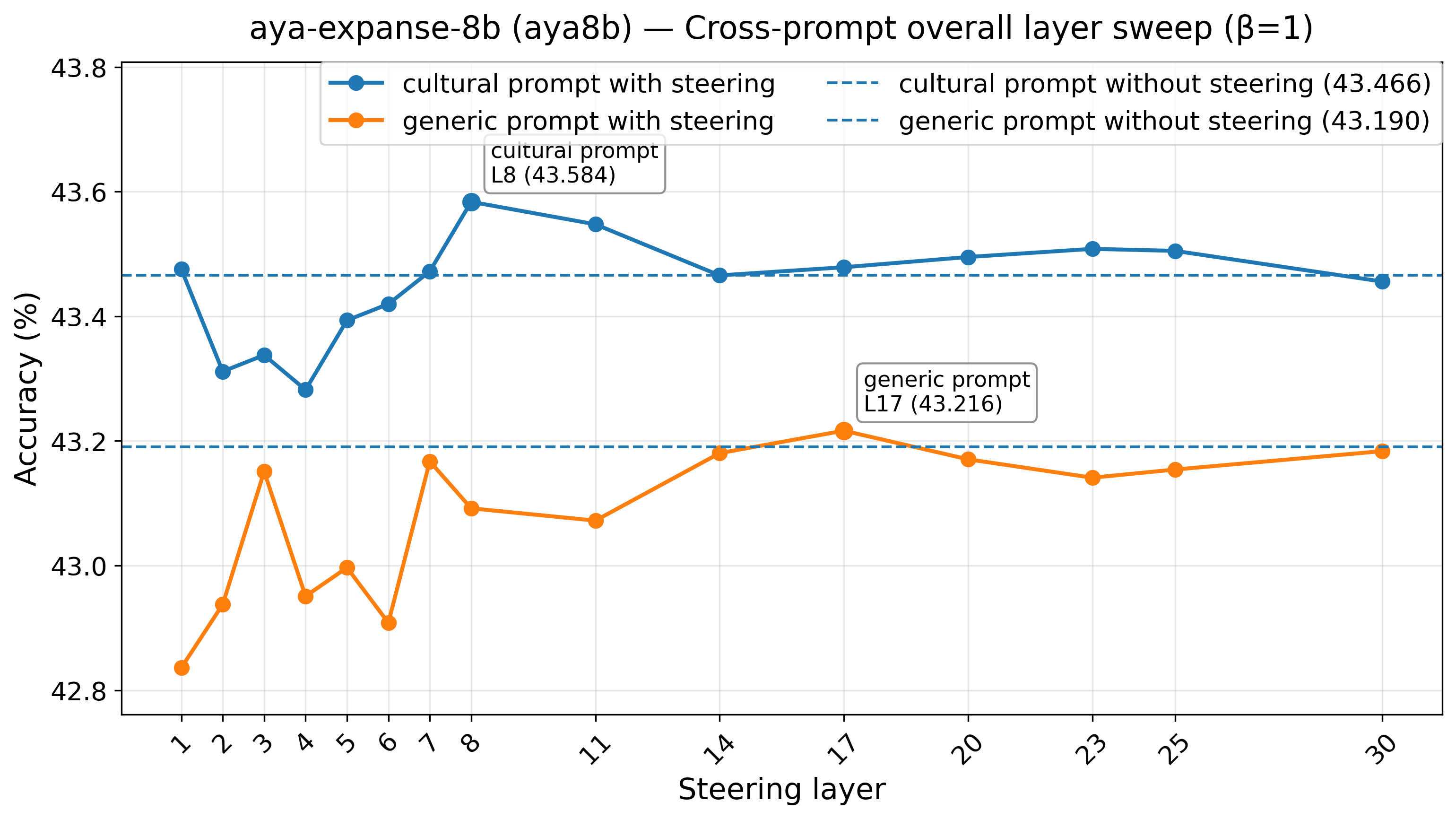}
    \caption{Post-hoc cross-prompt SAQ layer sweep for Aya Expanse 8B with $\beta=1$. The official submission uses the \textbf{cultural prompt}. Prompt choice affects both baseline accuracy and the optimal steering layer (here, Layer~8 for the cultural prompt and Layer~17 for the generic prompt).}
\end{figure}


\begin{figure}[t]
    \centering
    \includegraphics[width=0.9\linewidth]{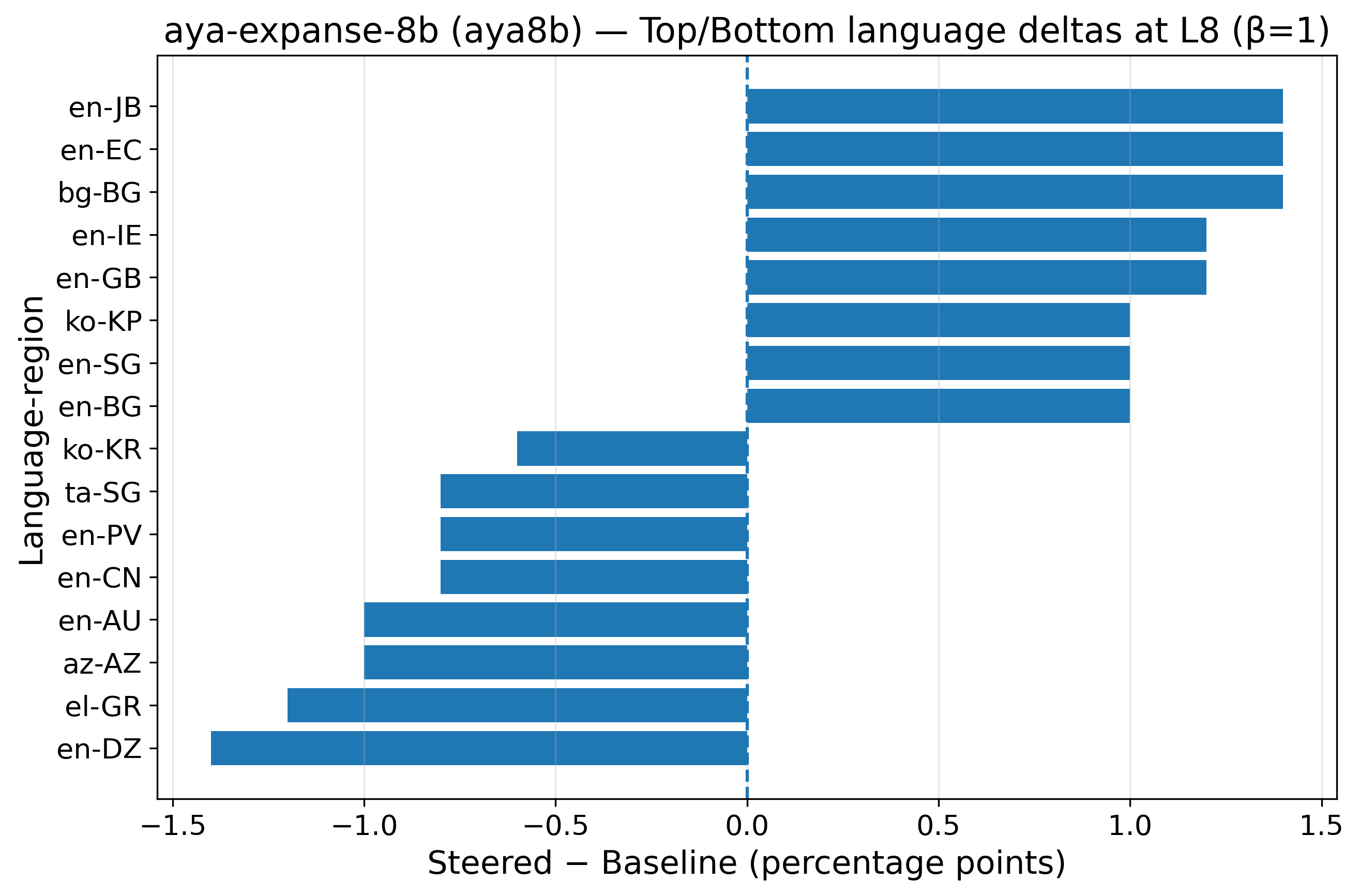}
    \caption{Top and bottom per-language SAQ accuracy changes (steered minus baseline, percentage points) for Aya Expanse 8B at Layer~8 with $\beta=1$ using the cultural prompt. Steering produces substantially different effects across language-region pairs, including both strong gains and degradations.}
\end{figure}


\newpage
% ...
\begin{figure}[t]
\subsection{Aya Expanse 32B}

    \centering
    \includegraphics[width=0.9\linewidth]{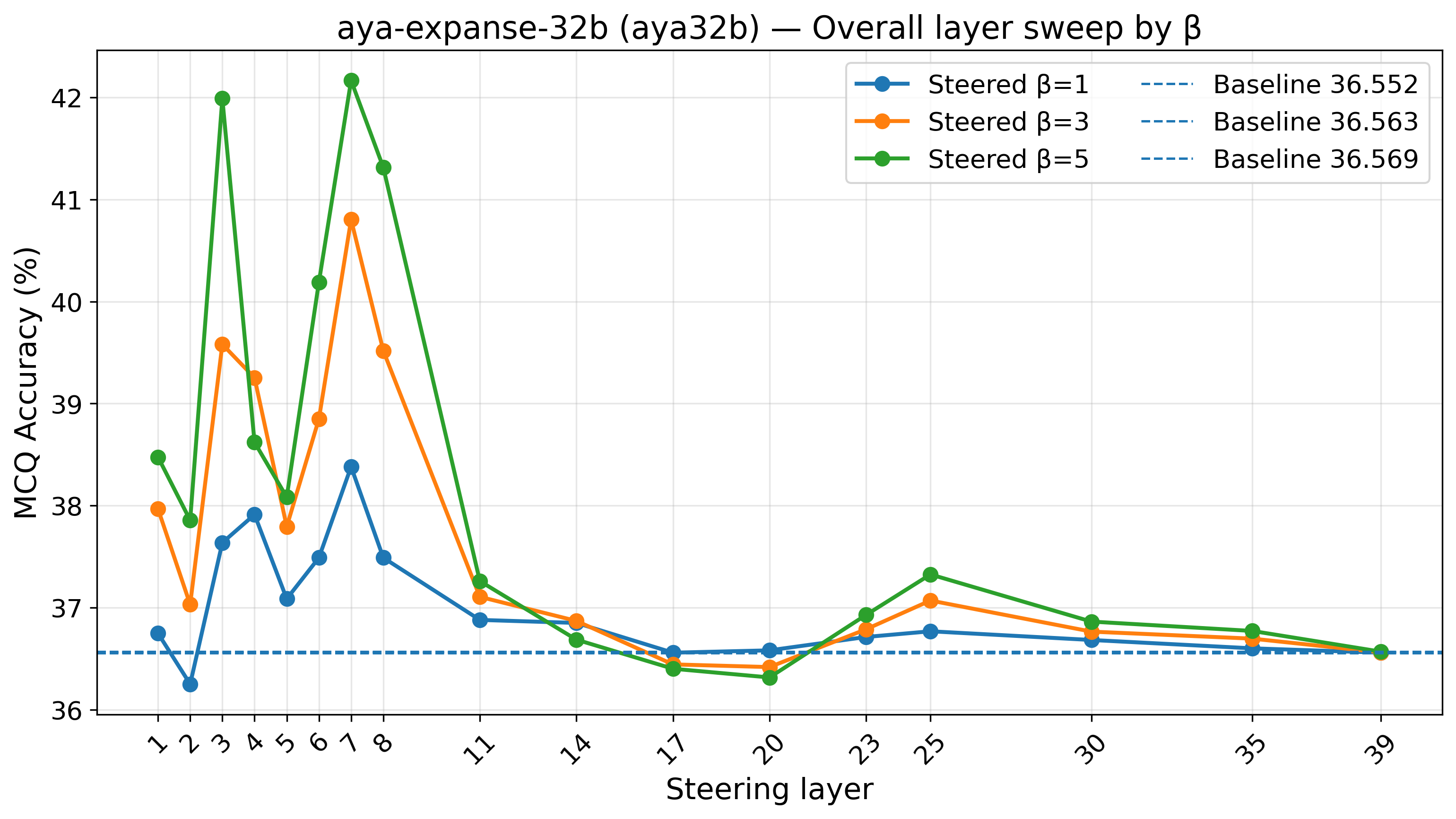}
    \caption{Post-hoc overall SAQ layer sweeps for Aya Expanse 32B under different steering strengths ($\beta \in \{1,3,5\}$). Large steering strengths can substantially improve performance in early layers and $\beta=5$ yields the best overall Acc in our experiments.}
\end{figure}


\begin{figure}[t]
    \centering
    \includegraphics[width=0.9\linewidth]{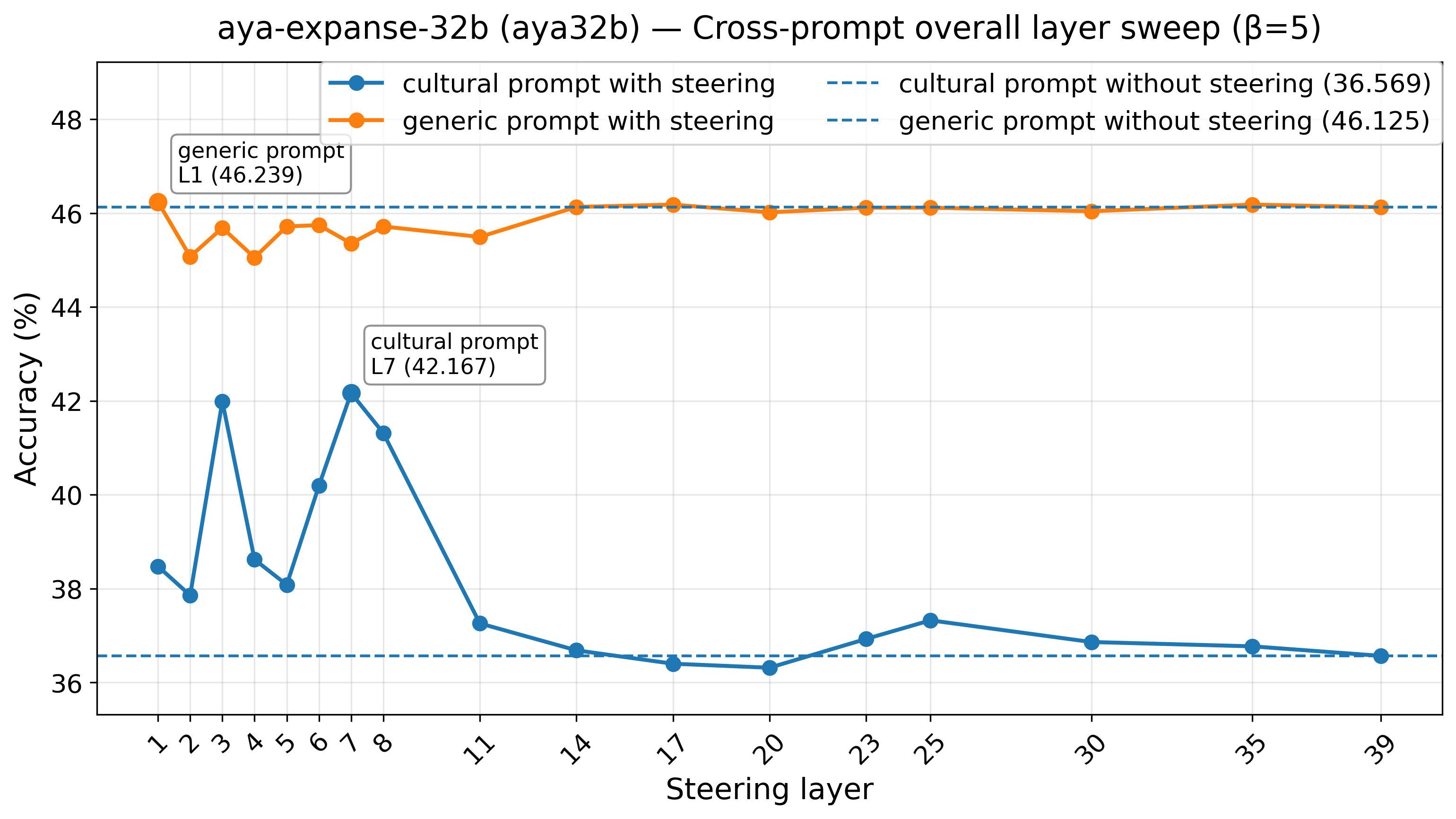}
    \caption{Post-hoc cross-prompt SAQ layer sweep for Aya Expanse 32B with $\beta=5$. The official submission uses the \textbf{cultural prompt}. Prompt choice affects both baseline accuracy and the optimal steering layer (here, Layer~7 for the cultural prompt and Layer~1 for the generic prompt).}
\end{figure}


\begin{figure}[t]
    \centering
    \includegraphics[width=0.9\linewidth]{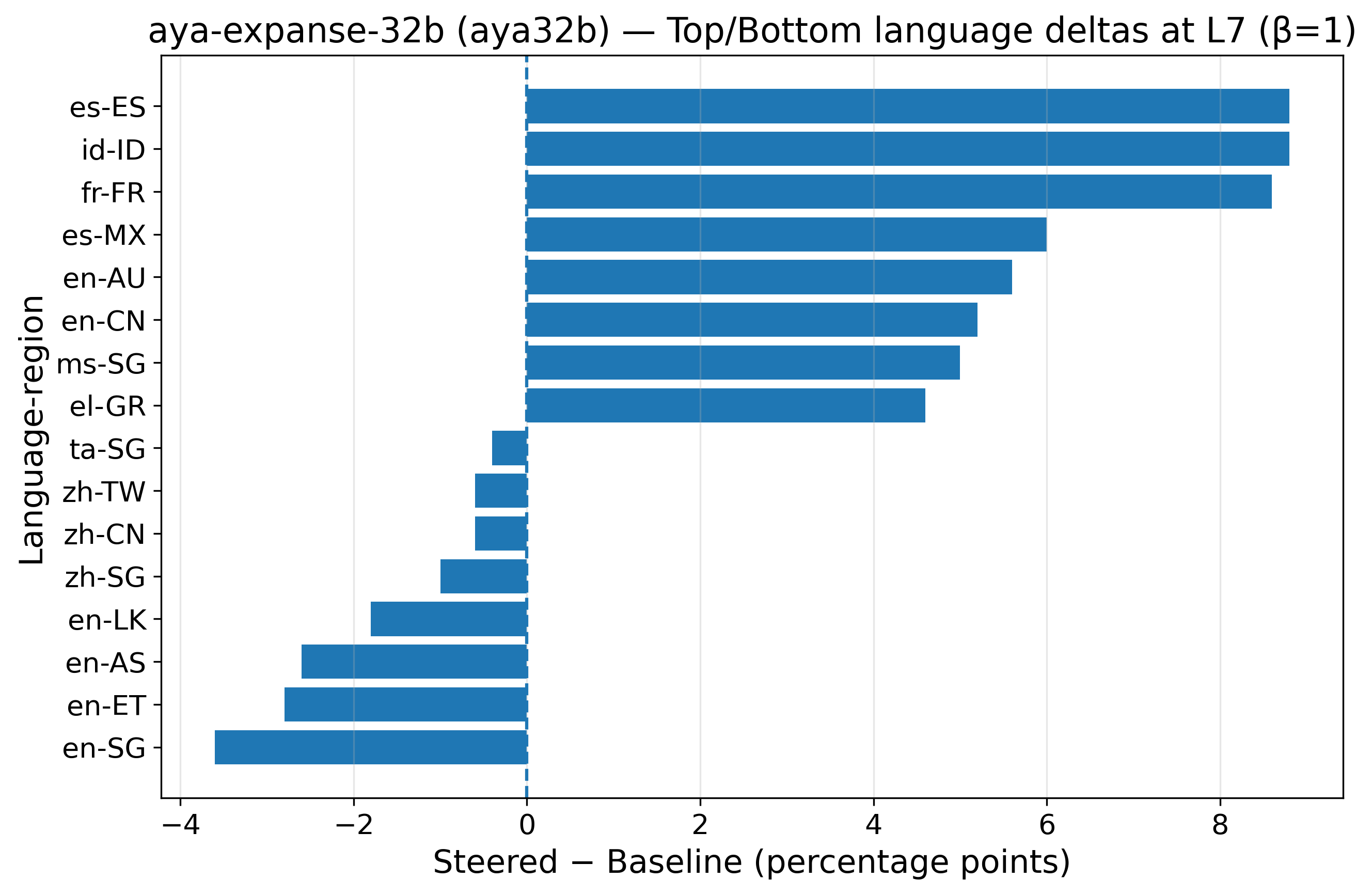}
    \caption{Top and bottom per-language SAQ accuracy changes (steered minus baseline, percentage points) for Aya Expanse 32B at Layer~7 with $\beta=1$ using the cultural prompt. Steering produces substantially different effects across language-region pairs, including both strong gains and degradations.}
\end{figure}


\newpage
% ...
\begin{figure}[t]
\subsection{Qwen3-8B}
    \centering
    \includegraphics[width=0.9\linewidth]{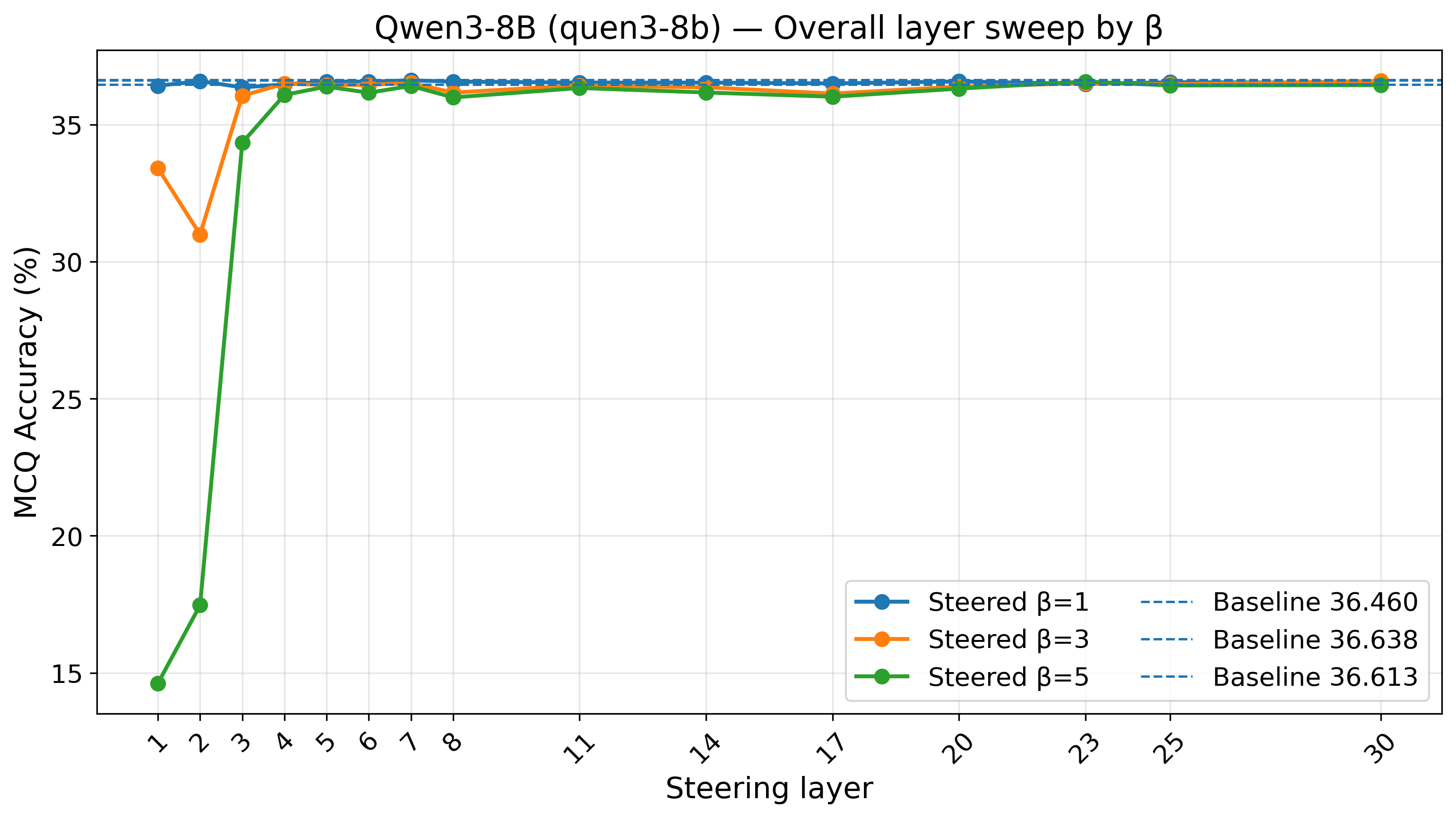}
    \caption{Post-hoc overall SAQ layer sweeps for Qwen3-8B under different steering strengths ($\beta \in \{1,3,5\}$). Large steering strengths can substantially degrade performance in early layers, while $\beta=1$ remains stable and yields the best overall trade-off in our experiments.}
\end{figure}


\begin{figure}[t]
    \centering
    \includegraphics[width=0.9\linewidth]{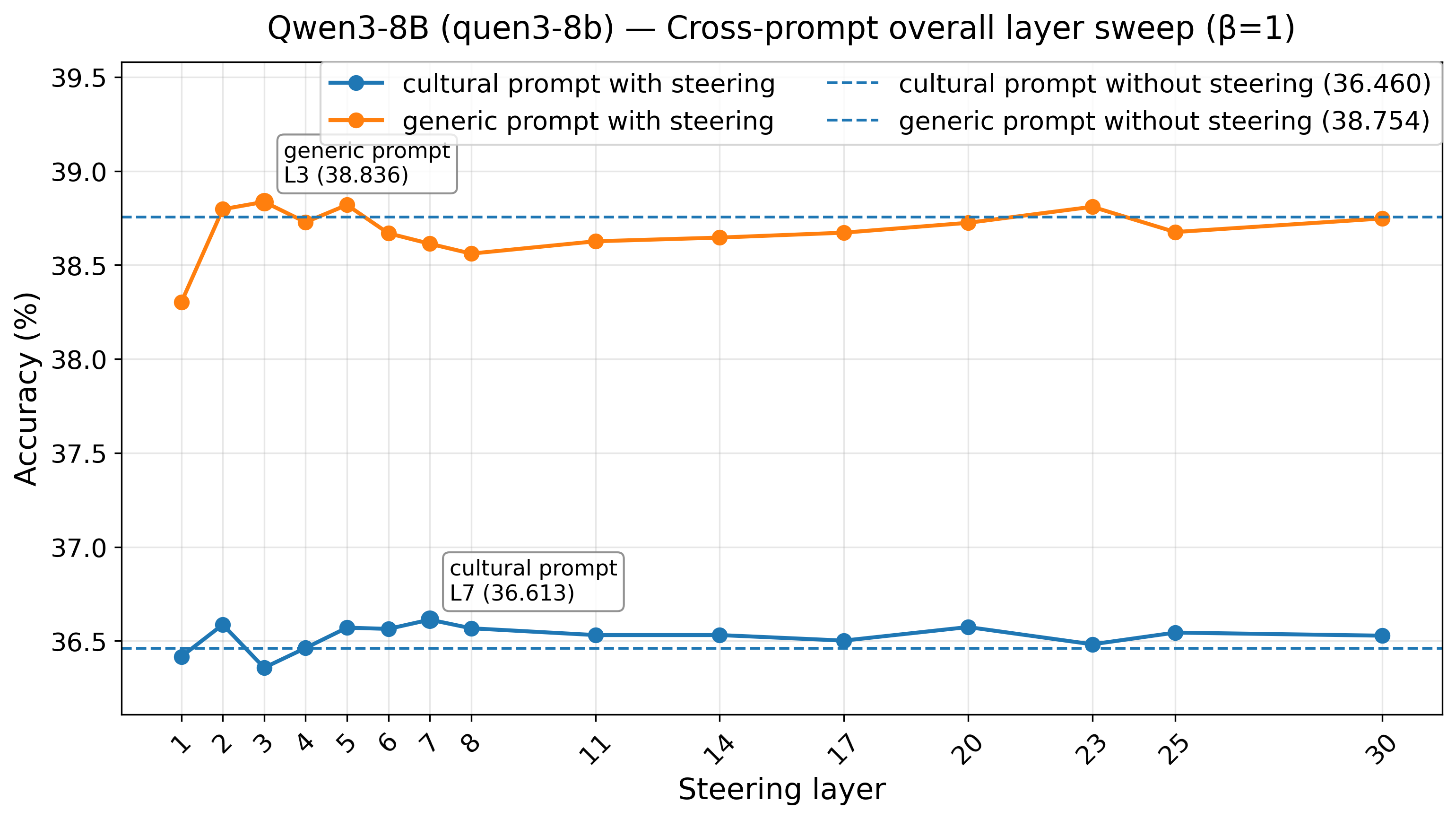}
    \caption{Post-hoc cross-prompt SAQ layer sweep for Qwen3-8B with $\beta=1$. The official submission uses the \textbf{cultural prompt}. Prompt choice affects both baseline accuracy and the optimal steering layer (here, Layer~3 for the cultural prompt and Layer~7 for the generic prompt).}
\end{figure}


\begin{figure}[t]
    \centering
    \includegraphics[width=0.9\linewidth]{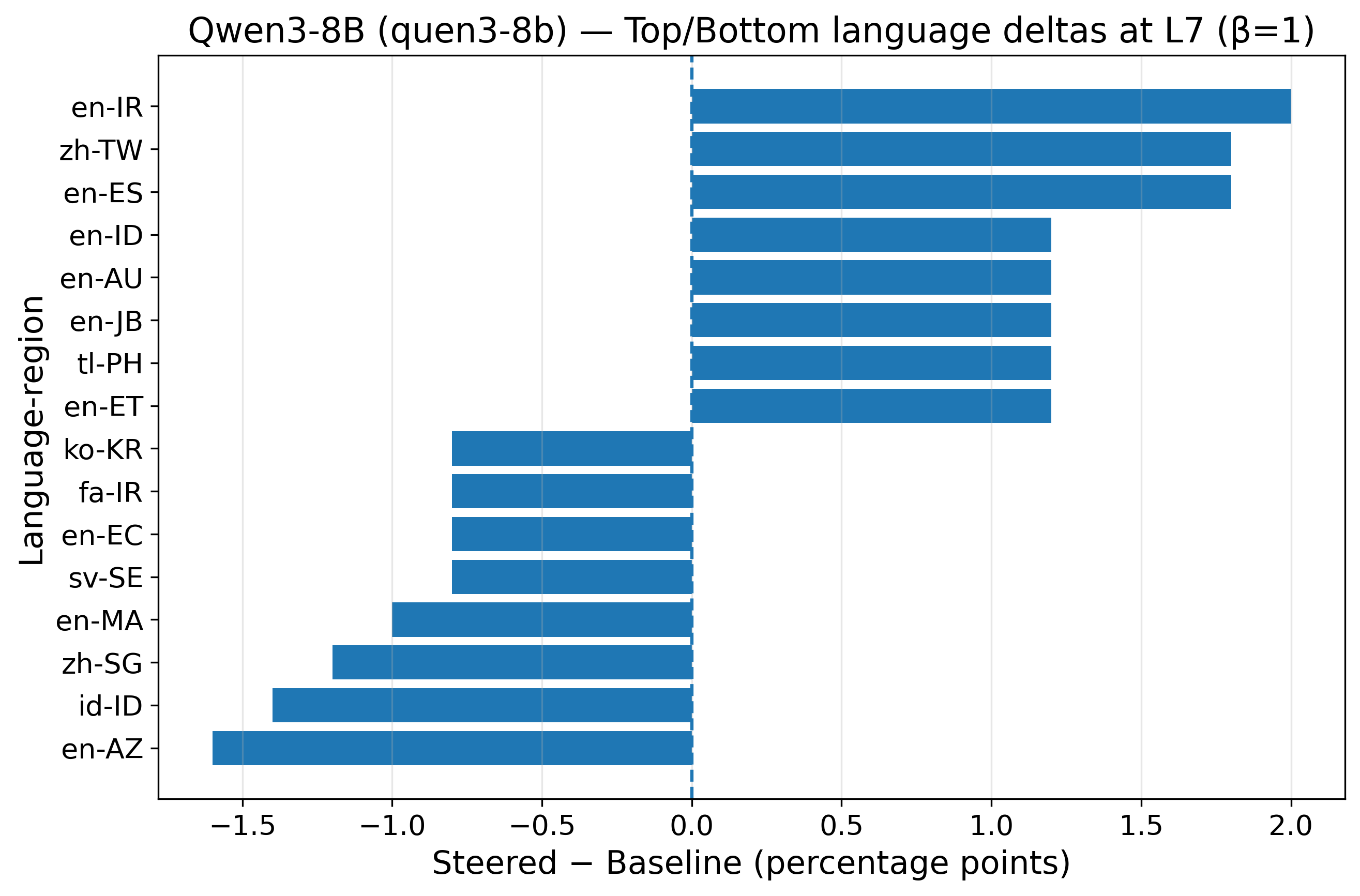}
    \caption{Top and bottom per-language SAQ accuracy changes (steered minus baseline, percentage points) for Qwen3-8B at Layer~7 with $\beta=1$ using the cultural prompt. Steering produces substantially different effects across language-region pairs, including both strong gains and degradations.}
\end{figure}


% ...
\newpage
% ...
\begin{figure}[t]
\subsection{Qwen3-32B}
    \centering
    \includegraphics[width=0.9\linewidth]{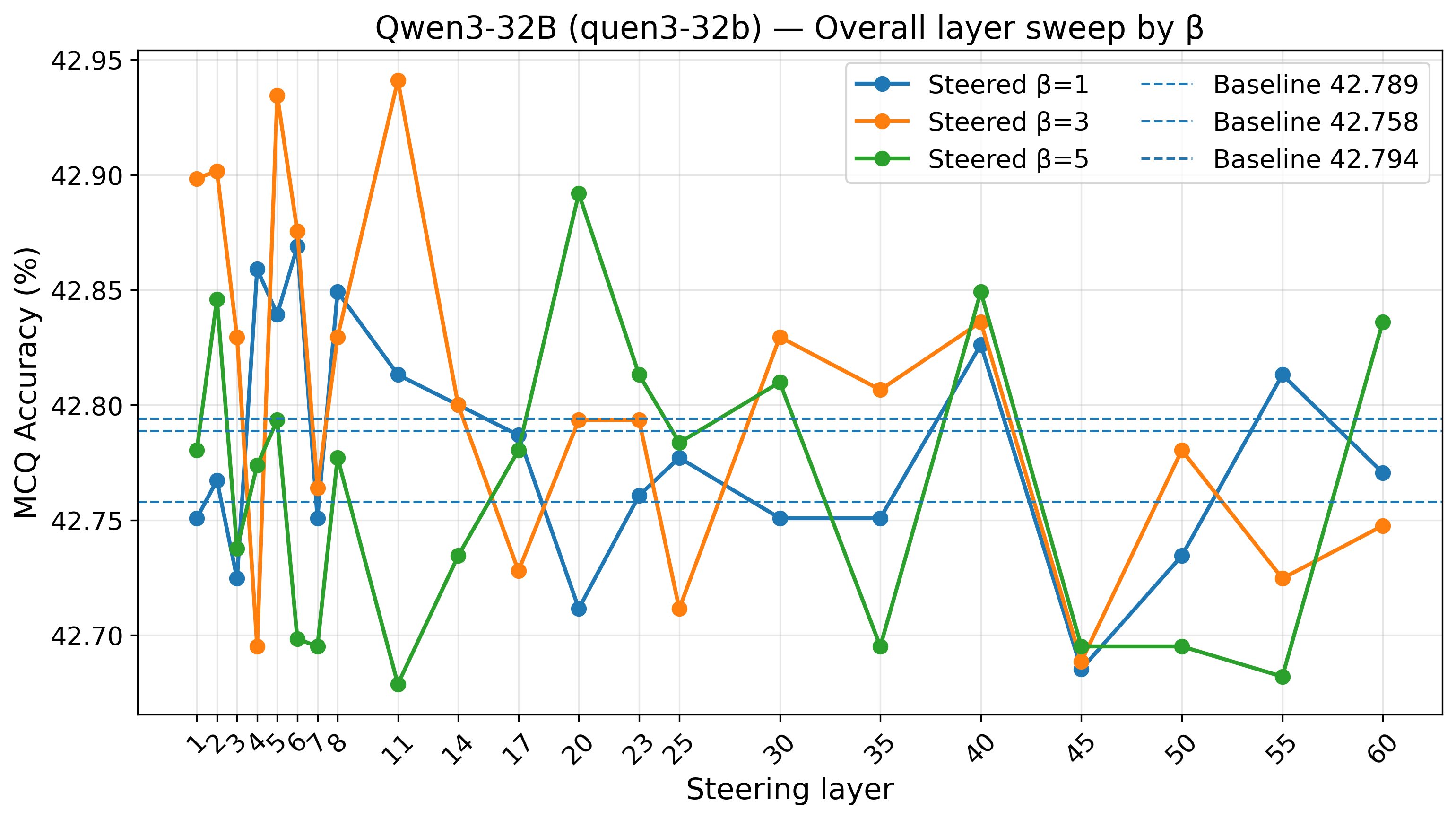}
    \caption{Post-hoc overall SAQ layer sweeps for Qwen3-32B under different steering strengths ($\beta \in \{1,3,5\}$). Steering strengths show unstable performance across all layers, while $\beta=5$ yields the best overall Acc in our experiments.}
\end{figure}


\begin{figure}[t]
    \centering
    \includegraphics[width=0.9\linewidth]{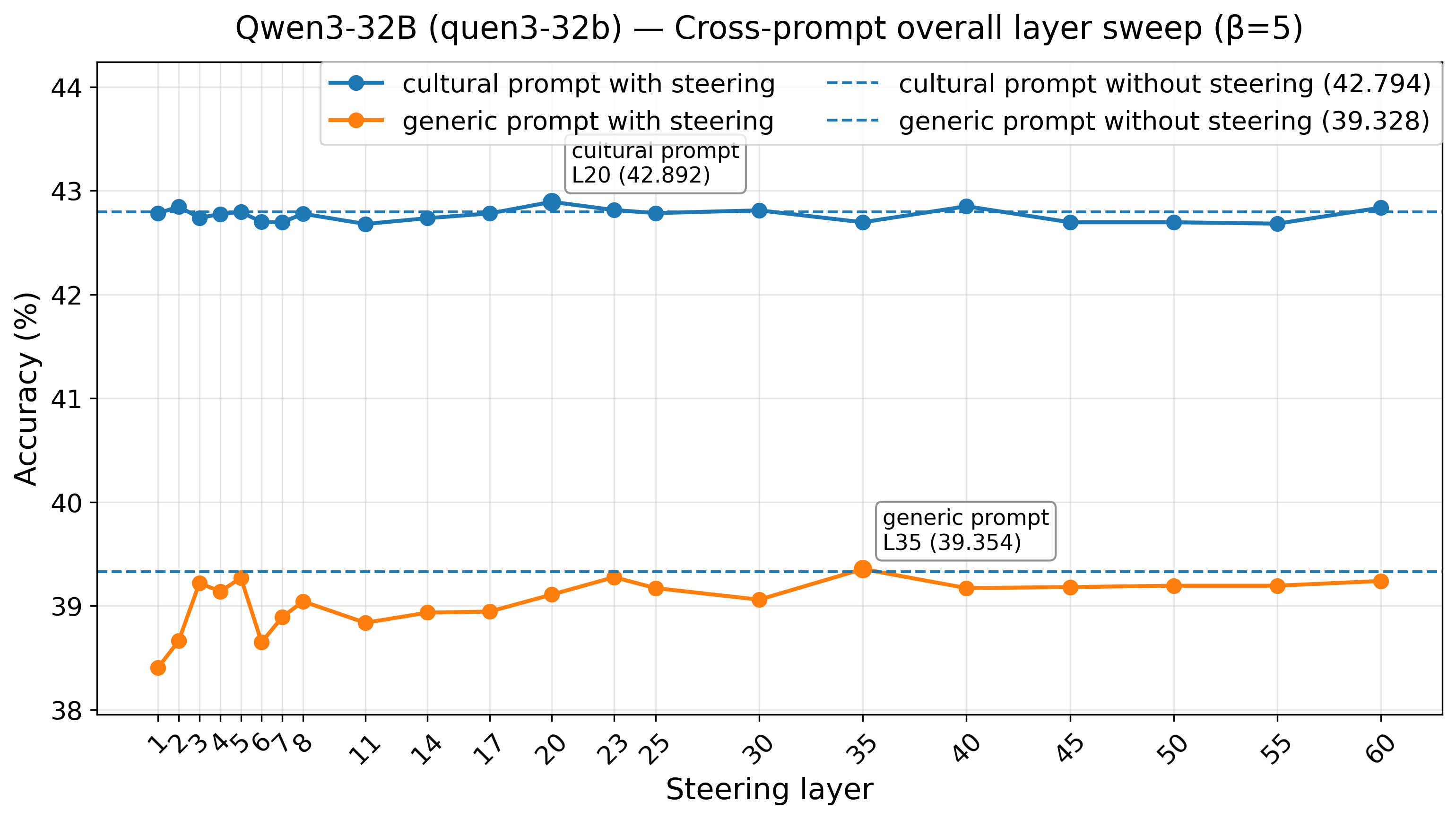}
    \caption{Post-hoc cross-prompt SAQ layer sweep for Qwen3-32B with $\beta=5$. The official submission uses the \textbf{cultural prompt}. Prompt choice affects both baseline accuracy and the optimal steering layer (here, Layer~20 for the cultural prompt and Layer~35 for the generic prompt).}
\end{figure}


\begin{figure}[t]
    \centering
    \includegraphics[width=0.9\linewidth]{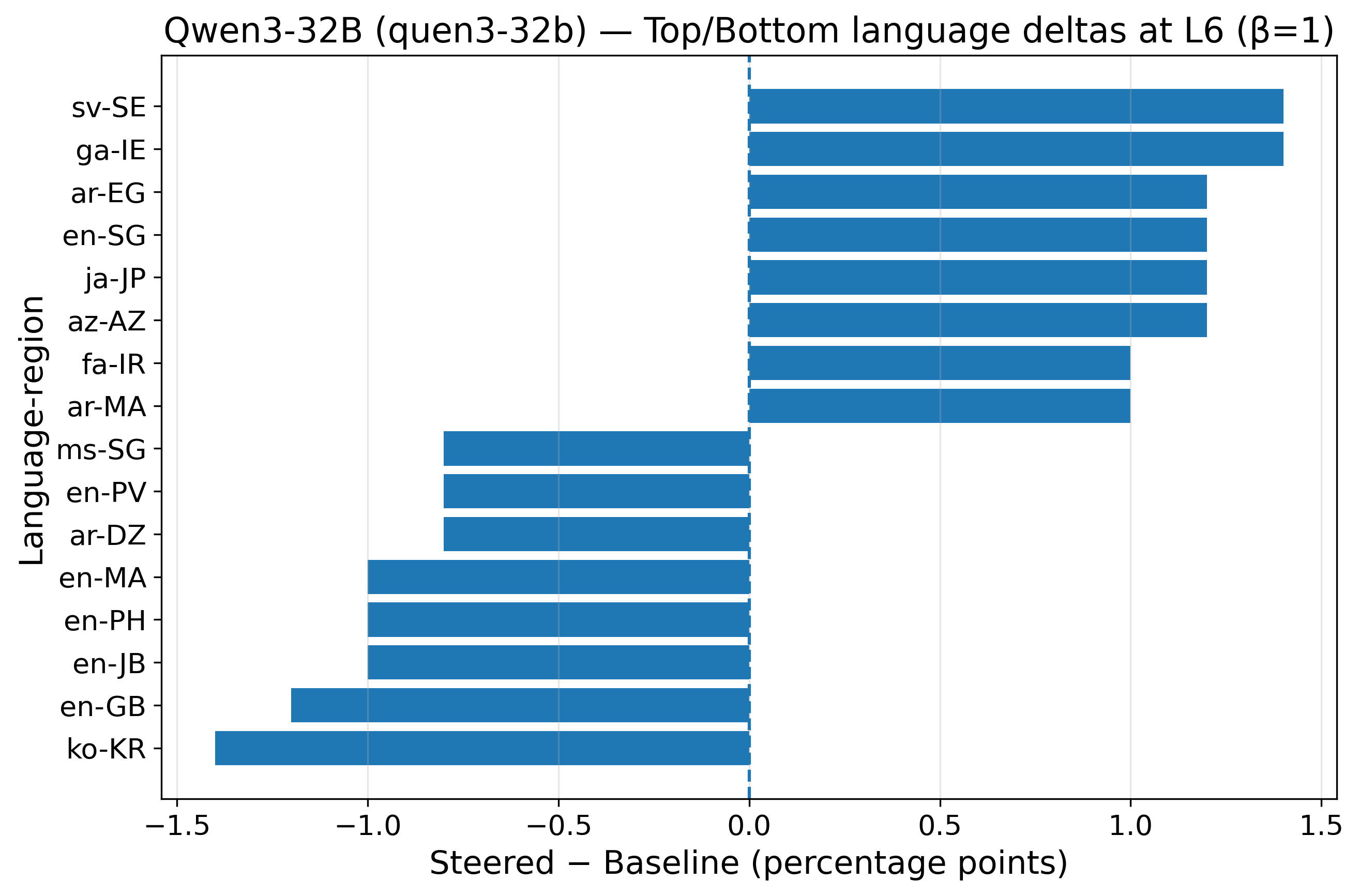}
    \caption{Top and bottom per-language SAQ accuracy changes (steered minus baseline, percentage points) for Qwen3-32B at Layer~6 with $\beta=1$ using the cultural prompt. Steering produces substantially different effects across language-region pairs, including both strong gains and degradations.}
\end{figure}


\clearpage

\section{Random- vs.\ Language-Vector Steering}
\label{app:rand-vs-lang}


This appendix summarizes the random-vector control used to test whether language-vector steering effects reflect language-specific structure or generic activation perturbations.

\paragraph{Setup.}
For each model, layer, prompt, and locale, we compare two unit-norm steering directions applied with the same intervention
\[
\tilde h^{(l)} = h^{(l)} + \beta u^{(l)}, \qquad \beta=1.
\]
The language direction is the FLORES DiffMean vector from \S3.1,
\[
v_\ell^{(l)}
=
\frac{1}{|D_\ell|}\sum_{x\in D_\ell}h^{(l)}(x)
-
\frac{1}{|D_{\neg \ell}|}\sum_{x\in D_{\neg \ell}}h^{(l)}(x),
\]
computed from token-mean post-normalization residual activations and then L2-normalized. The random control samples
\[
r_{\ell,k}^{(l)} \sim \mathcal{N}(0,I_d), \qquad
\hat r_{\ell,k}^{(l)} = r_{\ell,k}^{(l)} / \lVert r_{\ell,k}^{(l)} \rVert_2,
\]
with the same dimensionality and norm as the language vector. Within each draw $k$, the same random vector is reused for all locales sharing the corresponding FLORES language code, mirroring the language-vector setup.

\paragraph{Bootstrap.}
For the random baseline, we run four independent Gaussian draws and compute
\[
\Delta_{\mathrm{random},k}
=
\mathrm{acc}_{\mathrm{mod},k}
-
\mathrm{acc}_{\mathrm{base},k}.
\]
We then average $\Delta_{\mathrm{random},k}$ over draws and compare it to the single language-vector delta,
\[
\Delta_{\mathrm{language}}
=
\mathrm{acc}_{\mathrm{lang}}
-
\mathrm{acc}_{\mathrm{base}}.
\]
Both deltas are reported in percentage points. The comparison is matched at the same model, layer, prompt, locale, and steering strength.

\paragraph{Interpretation.}
The Qwen2.5-72B comparison in Figure~\ref{fig:rand-vs-lang-avg-scatter} shows that averaged random-vector effects remain concentrated near zero, while language-vector effects are somewhat more dispersed. This supports the main-text conclusion that language-vector steering is not simply equivalent to a generic Gaussian perturbation, but also does not yield reliably positive gains at the tested layers.


\end{document}